\documentclass[10pt,twocolumn,letterpaper]{article}

\usepackage{cvpr}
\usepackage{times}
\usepackage{epsfig}
\usepackage{graphicx}
\usepackage{amsmath}
\usepackage{amssymb}
\usepackage{rotating}
\usepackage{booktabs}
\usepackage{multirow}
\usepackage{subfig}
\usepackage[ruled,vlined]{algorithm2e}
\usepackage{amsthm}
\usepackage{bbm}
\usepackage{makecell}
\usepackage{mathtools}
\usepackage{url}
\usepackage[breaklinks=true,colorlinks,bookmarks=false]{hyperref}

\newcommand{\vect}[1]{\mathbf{#1}}
\newcommand{\matr}[1]{\mathbf{#1}}

\newcommand{\set}[1]{\mathcal{#1}}

\SetCommentSty{mycommfont}
\DeclareMathOperator*{\argmax}{arg\,max}

\cvprfinalcopy 

\def\cvprPaperID{****} 

\ifcvprfinal\fi
\begin{document}

\title{Weakly Supervised Semantic Point Cloud Segmentation: Towards 10$\times$ Fewer Labels}

\author{Xun Xu\thanks{now with A-STAR, Singapore. \url{http://xu-xun.com}}\qquad Gim Hee Lee\\
Department of Computer Science, National University of Singapore\\
{\tt\small alex.xun.xu@gmail.com, gimhee.lee@nus.edu.sg}
}

\maketitle

\begin{abstract}
Point cloud analysis has received much attention recently; and segmentation is one of the most important tasks. The success of existing approaches is attributed to deep network design and large amount of labelled training data, where the latter is assumed to be always available. However, obtaining 3d point cloud segmentation labels is often very costly in practice. In this work, we propose a weakly supervised point cloud segmentation approach which requires only a tiny fraction of points to be labelled in the training stage. This is made possible by learning gradient approximation and exploitation of additional spatial and color smoothness constraints. Experiments are done on three public datasets with different degrees of weak supervision. In particular, our proposed method can produce results that are close to and sometimes even better than its fully supervised counterpart
with 10$\times$ fewer labels. Our code is available at the project website\footnote{\url{https://github.com/alex-xun-xu/WeakSupPointCloudSeg}}. 
\end{abstract}
\vspace{-0.5cm}

\section{Introduction}

Recent developments in point cloud data research have witnessed the emergence of many supervised approaches \cite{Qi2017,qi2017pointnet++,wang2019dynamic,li2018pointcnn,wang2019graph}. Most efforts of current research are dedicated into two tasks: point cloud shape classification (a.k.a. shape recognition) and point cloud segmentation (a.k.a. semantic segmentation). For both tasks, the success of the state-of-the-art methods is attributed mostly to the deep learning architecture \cite{Qi2017} and the availability of large amount of labelled 3d point cloud data \cite{mo2019partnet,armeni20163d}. 
Although the community is still focused on pushing forward in the former direction, we believe the latter issue, i.e. data annotation, is an overlooked bottleneck. In particular, it is assumed that all points 
for the point cloud segmentation task
are provided with ground-truth labels, which is often in the range of 1k to 10k points for a 3d shape \cite{yi2016scalable,mo2019partnet}. The order of magnitude increases drastically to millions of points for a real indoor scene \cite{landrieu2018large}. As a result, very accurate labels for billions of points are needed in a dataset to train good segmentation models. Despite the developments of modern annotation toolkits \cite{mo2019partnet,armeni20163d} to facilitate large-scale annotation, exhaustive labelling is still prohibitively expensive for ever growing new datasets.


\begin{figure}
\centering
\includegraphics[width=1.02\linewidth]{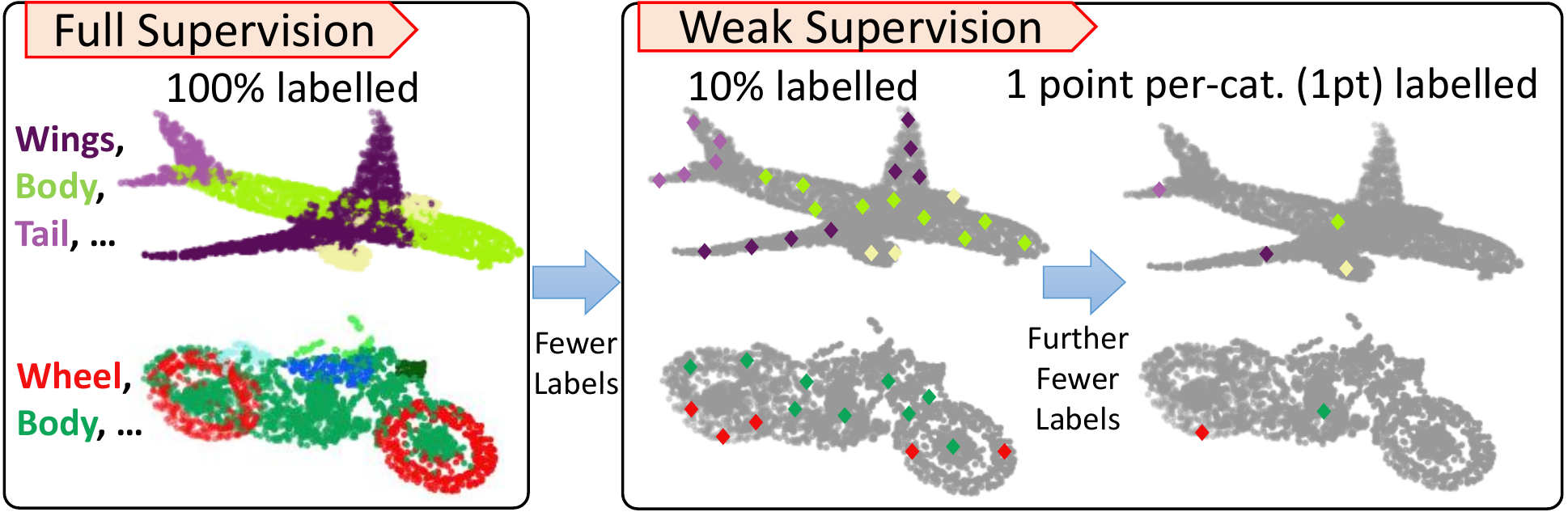}
\vspace{-0.5cm}
\caption{Illustration of the weak supervision concept in this work. Our approach achieves 
segmentation with only a fraction of labelled points.
.}\label{fig:WeakSupConcept}
\vspace{-0.5cm}
\end{figure}

In this work, we raise the question on whether it is possible to learn a point cloud segmentation model with only partially labelled points. And, if so, how many is enough for good segmentation. This problem is often referred to as weakly supervised learning in the literature \cite{Zhou2017review} as illustrated in Fig.~\ref{fig:WeakSupConcept}. To the best of our knowledge, there are only a handful of works which tried to address related problems \cite{guinard2017weakly,mei2019semantic}. In \cite{guinard2017weakly}, a non-parametric conditional random field classifier (CRF) is proposed to capture the geometric structure for weakly supervised segmentation. However, it casts the task into a pure structured optimization problem, and thus fail to capture the context, e.g. spatial and color cues. A method for semi-supervised 3D LiDAR data segmentation is proposed in
\cite{mei2019semantic}. It converts 3D points into a depth map with CNNs applied for feature learning, and the semi-supervised constraint is generated from the temporal consistency of the LiDAR scans. Consequently, it is not applicable to general 3D point cloud segmentation. 

To enable the weakly supervised segmentation with both strong contextual modelling ability and handling generic 3D point cloud data, we choose to build upon the state-of-the-art deep neural networks for learning point cloud feature embedding \cite{Qi2017,wang2019dynamic}. Given partially labelled point cloud data, we employ an incomplete supervision branch with softmax cross-entropy loss that penalizes only on labelled points. We observe that such simple strategy can succeed even at 10$\times$ fewer labels, i.e. only $10\%$ of the points are labelled. This is because the learning gradient of the incomplete supervision can be considered as a sampling approximation of the full supervision. In Sect.~\ref{sect:IncompleteSup}, we show our analysis that the approximated gradient converges to the true gradient in distribution, and the gap is subjected to a normal distribution with variance inversely proportional to the number of sampled points. As a result, the approximated gradient is close to the true gradient given enough labelled points. The analysis also gives an insight into choosing the best annotation strategy under fixed budget. We conclude that it is always better to extensively annotate more samples with fewer labelled points in each sample than to intensively label fewer samples with more (or fully) labelled points.  

\begin{figure}
\centering
\includegraphics[width=1.05\linewidth]{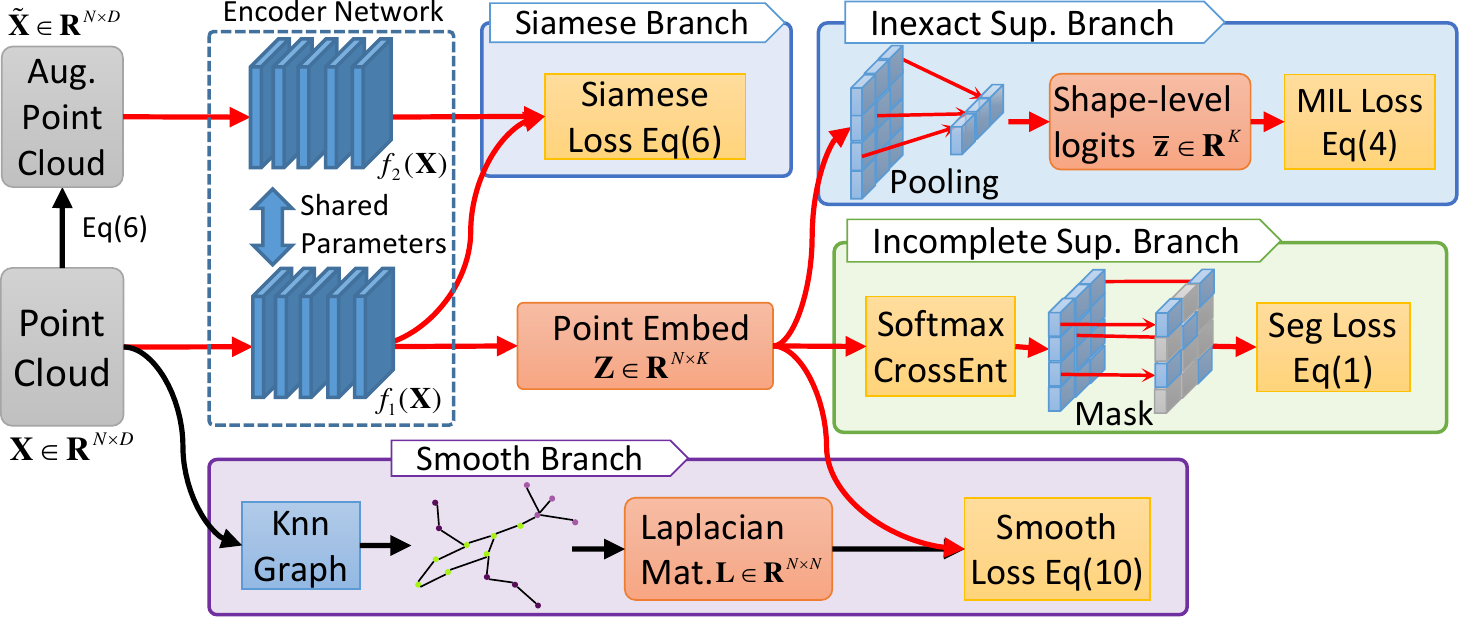}
\caption{Our network architecture for weakly supervised point cloud segmentation. Red lines indicate back propagation flow.
}\label{fig:Network}
\vspace{-0.5cm}
\end{figure}

As the above method imposes constraints only on the labelled points, we propose additional constraints to the unlabelled points in three orthogonal directions. First, we introduce an additional inexact supervision branch which defines a point cloud sample level cross entropy loss
in a similar way to multi-instance learning\cite{Zhou2015,ilse2018attention}. It aims to suppress the activation of any point with respect to the negative categories. Second, we introduce a Siamese self-supervision branch by augmenting the training sample with a random in-plane rotation and flipping, and then encourage the original and augmented point-wise predictions to be consistent. Finally, we make the observation that semantic parts/objects are often continuous in the spatial and color spaces. To this end, we propose a spatial and color smoothness constraint to encourage spatially adjacent points with similar color to have the same prediction. Such constraint can be further applied at inference stage by solving a soft constrained optimization that resembles label propagation on a graph \cite{zhu2003semi}. Our proposed network is illustrated in Fig.~\ref{fig:Network}.

\textbf{Our contributions} are fourfold. i) To the best of our knowledge, this is the first work to investigate weakly supervised point cloud segmentation within a deep learning context. ii) We give an explanation to the success of weak supervision and provide insight into annotation strategy under a fixed labelling budget. iii) We adopt three additional losses based on inexact supervision, self-supervision and spatial and color smoothness to further constrain unlabelled data. iv) Experiments are carried out on three public dataset which serve as benchmarks to encourage future research.


\section{Related Work}

Weakly supervised learning aims to use weaker annotations, often in the form of partially labelled dataset or samples. In this work, we follow the definition of weak supervision made by \cite{Zhou2017review}. More specifically, we are concern with two types of weak supervision: incomplete and inexact supervision. 

\vspace{-0.45cm}
\paragraph{Incomplete Supervision.} This is also referred to as semi-supervised learning in the literature \cite{zhu2003semi,belkin2006manifold,papandreou2015weakly,bearman2016s,Laine_ICLR17,Kipf_ICLR17,iscen2019label}. We interchangeably use semi-supervised, weakly supervised and weak supervision in this paper to refer to this type of supervision.
It is assumed that only partial instances are labelled, e.g. only a few images are labelled for the recognition task \cite{zhu2003semi,zhou2004learning,iscen2019label}, a few bounding boxes or pixels are labelled for the image segmentation task \cite{papandreou2015weakly,bearman2016s} or a few nodes are labelled for graph inference \cite{Kipf_ICLR17}. The success is often attributed to the exploitation of problem specific assumptions including graph manifold \cite{zhu2003semi,belkin2006manifold,Kipf_ICLR17}, spatial and color continuity \cite{papandreou2015weakly,bearman2016s}, etc. 
Another line of works are based on ensemble learning by introducing additional constraints such as consistency between original and altered data, e.g. the addition of noise \cite{rasmus2015semi}, rotation \cite{Laine_ICLR17} or adversarial training \cite{miyato2018virtual}. This has further inspired ensemble approaches \cite{tarvainen2017mean,radosavovic2018data} akin to data distillation. Up till now, most of these works emphasize on large-scale image data, while very limited works have addressed point cloud data. \cite{mei2019semantic} proposes a semi-supervised framework for point cloud segmentation. However, it does not directly learn from  point cloud data and the required annotation is quite large. \cite{guinard2017weakly} proposes to exploit the geometric homogeneity and formulated a CRF-like inference framework. Nonetheless, it is purely optimization-based, and thus fails to capture the spatial relation between semantic labels. In this work, we make use of the state-of-the-art deep neural networks, and incorporate additional spatial constraints to further regularize the model. Thus we take advantage of both spatial correlation provided by deep models and geometric priors.

\vspace{-0.45cm}
\paragraph{Inexact Supervision.} It is also referred as weakly annotation in the image segmentation community \cite{Kolesnikov16,shi2016weakly}. They aim to infer the per-pixel prediction from a per-image level annotation \cite{Kolesnikov16,shi2016weakly} for image segmentation tasks. The class activation map (CAM) \cite{Zhou2015} is proposed to highlight the attention of of CNN based on discriminative supervision. It is proven to be a good prior model for weakly supervised segmentation \cite{Kolesnikov16,Wang2018}. Inexact supervision is often complementary to incomplete supervision, and therefore, it is also used to improve semi-supervised image segmentation\cite{bearman2016s}. In this work, we introduce inexact supervision as a complement to incomplete supervision for the task of point cloud segmentation. 

\vspace{-0.45cm}
\paragraph{Point Cloud Analysis.} It is applied on 3D shapes and has received much attention in recent years. The PointNet \cite{Qi2017} is initially proposed to learn 3D point cloud feature through cascaded multi-layer perceptrons (mlps) for point cloud classification and segmentation. The following works \cite{qi2017pointnet++,wang2019dynamic,li2018pointcnn,wang2018deep,landrieu2018large} are subsequently proposed to exploit local geometry through local pooling or graph convolution. Among all tasks of point cloud analysis, semantic segmentation is of high importance due to its potential application in robotics and the existing works rely on learning classifiers at point-level \cite{Qi2017}. However, this paradigm requires exhaustive point-level labelling and does not scale well. To resolve this issue, we propose a weakly supervised approach that requires only a fraction of points to be labelled. We also note that \cite{te2018rgcnn} proposes to add spatial smoothness regularization to the training objective. \cite{choy20194d} proposes to refine prediction via CRF. Nevertheless, both works require full supervision, while our work is based on a more challenging weak supervision setting.

\section{Methodology}
\subsection{Point Cloud Encoder Network}
We formally denote the input point cloud data as $\{\matr{X}_b\}_{b=1\cdots B}$ with $B$ individual shapes (e.g. shape segmentation) or room blocks (e.g. indoor point cloud segmentation). Each sample $\matr{X}_b\in \mathcal{R}^{N\times F}$ consists of $N$ 3d points with the xyz coordinates and possibly additional features, e.g. RGB values. Each sample is further accompanied with per-point segmentation label $\vect{y}_b\in \{1,\cdots K\}^{N}$, e.g. fuselage, wing and engine of a plane. 
For  clarity 
, we denote the one-hot encoded label as $\matr{\hat{Y}}\in\{0,1\}^{B\times N\times K}$. 
A point cloud encoder network $f(\matr{X};\Theta)$ parameterized by $\Theta$ is employed to obtain the embedded point cloud features $\matr{Z}_b\in \set{R}^{N\times K}$. We note that the dimension of the embedding is the same as the number of segmentation categories.
The recent development on point cloud deep learning  \cite{Qi2017,qi2017pointnet++,li2018pointcnn} provides many candidate encoder networks, which are evaluated in the experiment section.

\subsection{Incomplete Supervision Branch}\label{sect:IncompleteSup}
We assume that only a few points in the point cloud samples $\{\matr{X}_b\}$ are labelled with the ground-truth. Specifically, we denote a binary mask as $\matr{M}\in\{0,1\}^{B\times N}$, which is 1 for a labelled point and 0 otherwise. 
Furthermore, we define a softmax cross-entropy loss on the labelled point as 
\begin{equation}
l_{seg} = -\frac{1}{C}\sum_b\sum_im_{bi}\sum_k \hat{y}_{bik}\log\frac{\exp(z_{bik})}{\sum_k \exp(z_{bik})},
\end{equation}
where $C=\sum_{b,i}m_{bi}=||\matr{M}||_1$ is the normalization variable.

\begin{figure}[b]
\includegraphics[width=1\linewidth]{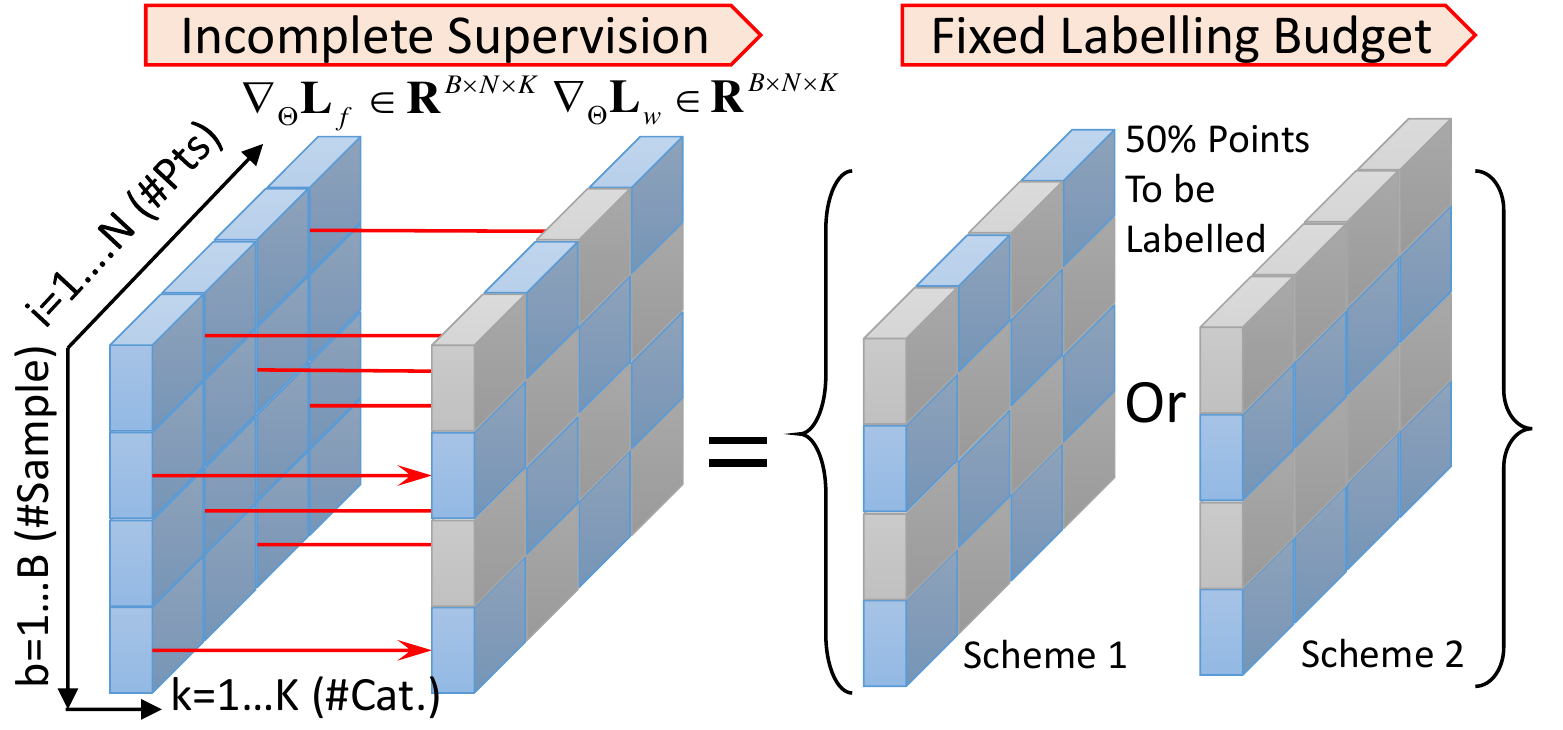}
\caption{Illustration of incomplete supervision and labeling strategies with fixed budget.}\label{fig:IncompleteSup}
\vspace{-0.5cm}
\end{figure}

\vspace{-0.4cm}
\paragraph{Discussion:} According to the experiments, we found that 
our method yields competitive results with as few as $10\%$ labelled points, i.e. $||\matr{M}||_1/(B\cdot N)=0.1$. The rationale is detailed in the following.
We first assume that two networks with similar weights -- one trained with full supervision and the other with weak supervision should produce similar results. 
Assuming that both networks start with an identical initialization, the higher similarity of the gradients at each step means a higher chance for the two networks to converge to similar results. Now, we write the gradients with full supervision $\nabla_\Theta l_f$ and weak supervision $\nabla_\Theta l_w$ as 
\begin{equation}\label{eq:Gradient}
\vspace{-0.2cm}
\begin{split}
\nabla_\Theta l_{f} &= \frac{1}{B\cdot N}\sum_b\sum_i\sum_k \nabla_\Theta l_{bik}, \quad \text{and} \\
\nabla_\Theta l_{w} &= \frac{1}{C}\sum_b\sum_im_{bi}\sum_k \nabla_\Theta l_{bik}, \\
\text{where}\quad  l_{bik}&=-\hat{y}_{bik}\text{log}\frac{\exp(z_{bik})}{\sum_k \exp(z_{bik})}.
\end{split}
\end{equation}
This relation is also illustrated in Fig.~\ref{fig:IncompleteSup}.

At each training step, the direction of the learning gradient is the mean of the gradients calculated with respect to each individual point. Suppose that $\nabla_\Theta l_{bik}$ is i.i.d. with expectation $E[\nabla_\Theta l_{bik}]=\mu$ and variance $Var[\nabla_\Theta l_{bik}]=\sigma^2$, and sampled mean (n samples) $S_n=mean(\nabla_\Theta l_{bik})$. We can easily verify that $E[\nabla_\Theta l_{bik}]=\nabla_\Theta l_f$ and $S_n=\nabla_\Theta l_w$ with $n=C=||\matr{M}||_1$. According to the Central Limit Theorem, we have the following convergence in distribution:
\begin{equation}
\vspace{-0.2cm}
\begin{split}
&\sqrt{n}(S_n-\mu)\xrightarrow[]{d}\set{N}(0,\sigma^2),\\
\Rightarrow&\sqrt{||\matr{M}||_1}(\nabla_\Theta l_w-\nabla_\Theta l_f)\xrightarrow[]{d}\set{N}(0,\sigma^2),\\
\Rightarrow&(\nabla_\Theta l_w-\nabla_\Theta l_f)\xrightarrow[]{d}\set{N}(0,\sigma^2/{||\matr{M}||_1}).
\end{split}
\end{equation}

This basically indicates that the difference between the gradient of full supervision and weak supervision is subjected to a normal distribution with variance $\sigma^2/||\matr{M}||_1$. Consequently, a sufficient number of labelled points, i.e. sufficiently large $||\matr{M}||_1$, is able to approximate $\nabla_\Theta l_f$ well with $\nabla_\Theta l_w$. Although the value of $\sigma$ is hard to estimate in advance, we empirically found that
our method yields results comparable to full supervision
with 10$\times$ fewer labelled points.

The analysis also provides additional insight into data annotation under a fixed budget. For example, with $50\%$ of the total points to be labelled as illustrated in Fig.~\ref{fig:IncompleteSup} (right):  
should we label 50\% of the points in each sample (Scheme 1) or label all the points in only 50\% of the samples (Scheme 2)? 
From the above analysis, it is apparent that Scheme 1 is better than Scheme 2 since it is closer to the i.i.d. assumption. This is further backed up by experiments in Sect.~\ref{sect:PtsVsSamp}.

\subsection{Inexact Supervision Branch}
In addition to the Incomplete Supervision Branch, a so-called inexact supervision
accompanies the annotation. Assuming each part has at least one labelled point, every training sample $\matr{X}_b$ is accompanied with an inexact label $\vect{\bar{y}}_b=\max_i\vect{\hat{y}}_{bi}$ simply by doing maxpooling over all points. Consequently, the inexact supervision branch is constructed in a similar fashion as multi-instance learning \cite{pathak2014fully,ilse2018attention}. The feature embedding $\matr{Z}_b$ is first globally max-pooled, i.e. $\vect{\bar{z}}_{b}=\max_i \vect{z}_{bi}$. 
We then introduce a loss for the inexact supervision branch. Since $\vect{\bar{z}}_b$ defines the logits on each category, the sigmoid cross entropy can be adopted as
\begin{equation}
\begin{split}
l_{mil} = &-\frac{1}{B\cdot K}\sum\limits_b\sum\limits_k \bar{y}_{bk}\log\frac{1}{1+\exp(-\bar{z}_{bk})} \\
&+ (1-\bar{y}_{bk})(\log(\frac{\exp(-\bar{z}_{bk})}{1+\exp(-\bar{z}_{bk})})).
\end{split}
\end{equation}
The rationale is that for those part categories that are absent from the sample, no points should be predicted with high logits. The incomplete supervision branch is only supervised on a tiny fraction of label points while the inexact supervision branch is supervised on the sample level with all points involved, so they are complementary to each other.


\subsection{Siamese Self-Supervision}
Despite the above two losses, majority of the unlabelled points are still not trained with any constraints. We believe additional constraints on those points can potentially further improve the results. To this end, we first introduce a Siamese self-supervision structure. We make the assumption that the prediction for any point is rotation and mirror flipping invariant. This assumption in particular holds true for 3D CAD shapes and indoor scenes with rotation in the XoY plane, e.g. the semantic label should not change with different view angle in a room. With this in mind, we design a Siamese network structure with two shared-parameter encoders $f_1(\matr{X})$ and $f_2(\matr{X})$. Then given a training sample $\matr{X}$, we apply a random transformation that consists of a random mirroring along the X and/or Y axes and an XoY plane rotation, i.e.
\vspace{-0.2cm}
\begin{equation}\label{eq:Tform}
\resizebox{0.90\linewidth}{!}{
$
\matr{R}=
\begin{bmatrix} 
cos\theta & -sin\theta & 0 \\
sin\theta & cos\theta & 0 \\
0 & 0 & 1
\end{bmatrix}\cdot
\begin{bmatrix} 
(2a-1)c & (2b-1)(1-c) & 0 \\
(2a-1)(1-c) & (2b-1)c & 0 \\
0 & 0 & 1
\end{bmatrix},
$
}
\end{equation}
where $\theta\sim \mathcal{U}(0,2\pi)$ (uniform distribution) and $a,b,c\sim\mathcal{B}(1,0.5)$ (Bernoulli distribution). Specifically, the first matrix controls the degree of rotation and the second matrix controls mirroring and X,Y swapping.
With the augmented sample denoted as $\tilde{\matr{X}}=\matr{X}\matr{R}^\top$, the rotation invariant constraint is turned into minimizing the divergence between the probabilistic predictions of $g(f_1(\matr{X}))$ and $g(f_2(\tilde{\matr{X}}))$, where $g(\cdot)$ is the softmax function. We use L2 distance to measure the divergence:
\begin{equation}
l_{sia}=\frac{1}{B\cdot N \cdot K}\sum_b||g(f_1(\matr{X}_b)) - g(f_2(\tilde{\matr{X}}_b))||_F^2,
\end{equation}
and empirically found it to be better than KL-Divergence.

\subsection{Spatial \& Color Smoothness Constraint}\label{sect:Smooth}
Semantic labels for 3D shape or scenes are usually smooth in both spatial and color spaces. Although they can be included by the state-of-the-art convolution networks \cite{wang2019graph}, explicit constraints are more beneficial in our context of weak supervision when the embedding of large amount of unlabelled points are not well constrained by the segmentation loss. Consequently, we introduce additional constraints at both training and inference stages.

\vspace{-0.4cm}
\paragraph{Spatial \& Color Manifold.} A manifold can be defined on the point cloud to account for the local geometry and color by a graph. We denote the 3D coordinate channels and RGB channels, if any, as $\matr{X^{xyz}}$ and $\matr{X^{rgb}}$, respectively. To construct a graph for the manifold, we first compute the pairwise distance $\matr{P}_c\in\set{R}^{N\times N}$ for channel $c$ (xyz or rgb) as $p_{ij}^c = ||\vect{x}_{i}^c-\vect{x}_{j}^c||_2,~\forall i,j\in \{1,\cdots N\}$.
A k-nn graph can be then constructed by searching for the k nearest neighbors $NN_k(\vect{x})$ of each point, and the corresponding weight matrix $\matr{W}^c\in\set{R}^{N\times N}$ is written as
\begin{equation}
\resizebox{0.90\linewidth}{!}{
$
w_{ij}^c=\left\{
\begin{array}{ll}
\text{exp}(-p_{ij}^c/\eta),\quad j\in NN_k(\vect{x}_i)\\
0, \quad \text{otherwise}
\end{array}, \forall i,j\in \{1,\cdots N\}.
\right.
$
}
\end{equation}
We take the sum of both weight matrices as $w_{ij}=w_{ij}^{xyz} + w_{ij}^{rgb}~\forall i,j$ to produce a more reliable manifold
when both xyz and rgb channels are available. This is reasonable since the xyz channel blurs the boundary and the rgb channel links faraway points, respectively. In case the manifold constructed on spatial distance and color contradicts the labelled ground-truth, we add additional must-link and must-not-link constraints \cite{wang2014constrained} to $\matr{W}$  to strengthen the compliance to known annotations, i.e.  
\begin{equation}\label{eq:MustLink}
w_{ij} = \left\{
\begin{array}{ll}
1,\quad m_{i},m_j=1, y_i = y_j\\
-1, \quad m_{i},m_j=1, y_i \neq y_j
\end{array}.
\right.
\end{equation}
 We further write the Laplacian matrix \cite{belkin2006manifold} as $\matr{L}=\matr{D}-\matr{W}$
with the degree matrix denoted as $\matr{D}=diag(\vect{d})$ \cite{von2007tutorial} and $d_i=\sum_j w_{ij}, \forall i\in \{1\cdots N\}$.

\vspace{-0.4cm}
\paragraph{Training Stage.} 
We introduce a manifold regularizer \cite{belkin2006manifold} to encourage the feature embedding of each point to comply with the manifold obtained previously. More specifically, the prediction $f(\vect{x}_i)$ should stay close to $f(\vect{x}_j)$ if $w_{ij}$ indicates high and stay unconstrained otherwise. Thus the regularizer is given by 
\begin{equation}
\resizebox{0.85\linewidth}{!}{
$
\begin{split}
l_{smo}&=\frac{1}{||\matr{W}||_0}\sum_i\sum_j w_{ij}||f(\vect{x}_i)-f(\vect{x}_j)||_2^2\\
&=\frac{2}{||\matr{W}||_0}(\sum_i d_{i}f(\vect{x}_i)^\top f(\vect{x}_i) - \sum_i \sum_j w_{ij} f(\vect{x}_i)^\top f(\vect{x}_j))\\
&=\frac{2}{||\matr{W}||_0}(tr(\matr{Z}^\top\matr{D}\matr{Z}) - tr(\matr{Z}^\top\matr{W}\matr{Z}))=\frac{2}{||\matr{W}||_0}tr(\matr{Z}^\top\matr{L}\matr{Z}),
\end{split}
$
}
\end{equation}
where $\matr{Z}$ is the prediction of all points.

\vspace{-0.4cm}
\paragraph{Inference Stage.} It is well known in image segmentation that the predictions of a CNN do not consider the boundaries well \cite{chen2017deeplab,Kolesnikov16} and CRF is often employed to refine the raw predictions. In weakly supervised point cloud segmentation, this issue exacerbates due to limited labels. To mitigate this problem, we introduce a semi-supervised label propagation procedure\cite{zhu2003semi} to refine the predictions. Specifically, the refined predictions $\matr{\tilde{Z}}$ should comply with the spatial and color manifold defined by the Laplacian $\matr{L}$, and at the same time should not deviate too much from the network predictions $\matr{Z}$. We write the objective as 
\begin{equation}
\begin{split}
&\min_{\{\vect{\tilde{z}}\}} \sum_i\sum_j w_{ij}||\vect{\tilde{z}}_i-\vect{\tilde{z}}_j||_2^2 + \gamma \sum_i ||\vect{\tilde{z}}_i-\vect{z}_i||_2^2, \\
\implies &\min_{\matr{\tilde{Z}}} tr(\matr{\tilde{Z}}^\top\matr{L}\matr{\tilde{Z}}) + \gamma ||\matr{\tilde{Z}} - \matr{Z}||_F^2.
\end{split} \raisetag{20pt}
\end{equation}
A closed-form solution exists for the above optimization \cite{zhu2003semi} and the final prediction for each point is simply obtained via
\begin{equation}\label{eq:LPClosedForm}
\begin{split}
&\tilde{y}_i=\argmax_k {\tilde{z}}_{ik},~\forall i\in\{1,\cdots N\}, \quad \text{where}\\
&\matr{\tilde{Z}}=\gamma(\gamma \matr{I} + \matr{L})^{-1}\matr{Z}.
\end{split}
\end{equation}

\subsection{Training}
The final training objective is the combination of all the above objectives, i.e. $l_{total}=l_{seg}+\lambda_1 l_{mil}+\lambda_2 l_{sia}+\lambda_3 l_{smo}$. We empirically set $\lambda_1,\lambda_2,\lambda_3=1$. The k-nn graph is selected as $k=10$, $\eta=1e3$, and $\gamma$ in Eq.~(\ref{eq:LPClosedForm}) is chosen to be 1. For efficient training, we first train the network with segmentation loss $l_{seg}$ only for 100 epochs. Then the total loss $l_{total}$ is trained for another 100 epochs. The default learning rate decay and batchnorm decay are preserved during the trainings
of different encoder networks. The initial learning rate is fixed at $1e-3$ for all experiments and the batchsize varies from 5 to 32 for different dataset bounded by the GPU memory size. Our algorithm is summarized in Algo.~\ref{alg:WeakSupSeg}. 

\begin{algorithm}[hb]
\SetKwData{Left}{left}\SetKwData{This}{this}\SetKwData{Up}{up}
\SetKwFunction{Concatenate}{Concatenate}\SetKwFunction{Kmeans}{K-means}
\SetKwInOut{Input}{input}\SetKwInOut{Output}{output}
\Input{\small{Point Cloud $\{\matr{X}_b\in \set{R}^{N\times D}\}$, Labels $\{\matr{y}_b\in\set{Z}^{N}\}$}}
\Output{\small{Segmentation Predictions $\{\vect{\tilde{y}}_b\in\set{Z}^{N}\}$}}
\tcc{Training Stage:}
\For{$epoch \leftarrow 1$ \KwTo $100$}
{
	Train One Epoch: $\Theta=\Theta - \alpha\nabla_\Theta l_{seg}|_{\{\matr{X}_b\},\{\vect{y}_b\}}$\;
}
\For{$epoch \leftarrow 1$ \KwTo $100$}
{
\tcp{Siamese Network}
Sample $\phi\sim \mathcal{U}(0,2\pi)$ and $a,b,c\sim\mathcal{B}(1,0.5)$\;
Calculate $\matr{R}$ according to Eq.~(\ref{eq:Tform})\;
Generate augmented sample $\matr{\tilde{X}}=\matr{X}\matr{R}^\top$\;
\tcp{Manifold Regularization}
Construct Laplacian $\matr{L}$ according to Sect.~\ref{sect:Smooth}\;
Train one epoch: $\Theta=\Theta - \alpha\nabla_\Theta l_{total}|_{\{\matr{X}_b\},\{\matr{\tilde{X}}_b\},\{\vect{y}_b\}}$\;
}
\tcc{Inference Stage:}
Forward pass $\matr{Z}_b=f(\matr{\tilde{X}}_b;\Theta)$\;
Obtain predictions $\{\vect{\tilde{y}}_b\}$ via Eq.~(\ref{eq:LPClosedForm})\;
\caption{\small{Weakly Supervised Point Cloud Segmentation}\label{alg:WeakSupSeg}}
\end{algorithm}

\section{Experiment}

\subsection{Dataset}
We conduct experiments of our weakly supervised segmentation model on three benchmark point cloud datasets. \textbf{ShapeNet} \cite{yi2016scalable} is a CAD model dataset with 16,881 shapes from 16 categories, each annotated with 50 parts. It is widely used as the benchmark for classification and segmentation evaluation.
We propose a weakly supervised setting. For each training sample we randomly select a subset of points from each part to be labelled. We use the default evaluation protocol for comparison. \textbf{PartNet} \cite{mo2019partnet} is proposed for more fine-grained point cloud learning. It consists of 24 unique shape categories with a total of 26,671 shapes. For the semantic segmentation task, it involves three levels of fine-grained annotation and we choose to evaluate at level 1. The incomplete weakly supervised setting is created in a similar way to ShapeNet, and we follow the original evaluation protocol. \textbf{S3DIS} \cite{armeni20163d} is proposed for indoor scene understanding. It consists of 6 areas each covering several rooms. Each room is scanned with RGBD sensors and is represented by point cloud with xyz coordinate and RGB value. For weakly supervised setting, we assume a subset of points are uniformly labelled within each room. The evaluation protocol on Area 5 as holdout is adopted.  

%


\subsection{Weakly Supervised Segmentation}

Two weakly supervision settings are studied. i) 1 point label (1pt), we assume there is only 1 point within each category labelled with ground-truth. Less than $0.8\%$ of total points are labelled for ShapeNet under the 1pt scheme. For S3DIS, the total labelled points is less than $0.2\%$. ii) 10 percentage label ($10\%$), we uniformly label $10\%$ of all points for each training sample. 

\vspace{-0.4cm}
\paragraph{Encoder Network.} We choose DGCNN \cite{wang2019dynamic} with default parameters as our encoder network due to its superior performance on benchmark shape segmentation and high training efficiency. However, as we point out in Sect.~\ref{sect:EncoderNet}, the proposed weakly supervised methods are compatible with alternative encoder networks. 

\vspace{-0.4cm}
\paragraph{Comparisons.} We compare against 3 sub-categories of methods. i) Fully supervised approaches (Ful.Sup.), including the state-of-the-art networks for point cloud segmentation. These methods serve as the upper bound of weakly supervised approaches. ii) Weakly supervised approaches (Weak Sup.), we implemented several generic weakly supervised methods and adapt them to point cloud segmentation tasks. In particular, the following methods are compared. 
The $\Pi$ model \cite{Laine_ICLR17} proposed to supervise on original input and the augmented input, but without the incomplete supervision on the augmented input. The mean teacher (MT) \cite{tarvainen2017mean} model employs a temporal ensemble for semi-supervised learning. The baseline method is implemented with only the segmentation loss $l_{seg}$ and DGCNN as encoder. Our final approach (Ours) is trained with the multi-task total loss $l_{total}$ with label propagation in the inference stage. iii) Unsupervised approaches, these methods do not rely on any annotations but instead directly infer clusters from the spatial and color affinities. Specifically, we experiment with Kmeans and normalized cut spectral clustering\cite{shi2000normalized} (Ncut). Both methods are provided with ground-truth number of parts.

\vspace{-0.4cm}
\paragraph{Evaluation.} For all datasets, we calculate the mean Intersect over Union (mIoU) for each test sample, and report the average mIoU over all samples (SampAvg) and all categories (CatAvg). For unsupervised methods, we find a best permutation between the prediction and ground-truth and then calculate the same mIoU metrics.

\vspace{-0.4cm}
\paragraph{ShapeNet.} We present the results in Tab.~\ref{tab:ShapeNet}, where we make the following observations. Firstly, the weak supervision model produces very competitive results
with only 1 labelled point per part category. The gap between full supervision and 1 point weak supervision is less than $12\%$. Secondly, we observe consistent improvement in the performance of segmentation with more labelled point from 1pt to 10$\%$. Interestingly, the weak supervision model is comparable to full supervision even
with 10$\%$ labelled points. Lastly, our proposed method that combines multiple losses and label propagation improves upon the baseline consistently, and outperforms alternative generic semi-supervised learning approaches and unsupervised clustering methods.
 
\begin{table*}[htbp]
  \centering
 \caption{\footnotesize{mIoU ($\%$) evaluation on ShapeNet dataset. The fully supervision (Ful. Sup.) methods are trained on $100\%$ labelled points. Three levels of weak supervisions (1pt, 1$\%$ and $100\%$) are compared. Ours method consists of DGCNN as encoder net, MIL branch, Siamese branch, Smooth branch and Inference label propagation.}}\vspace{-0.3cm}
   \setlength\tabcolsep{2pt} 
  \resizebox{0.99\linewidth}{!}{
      \begin{tabular}{cclcccccccccccccccccc}
    \toprule
    \multicolumn{2}{c}{Setting} & \multicolumn{1}{c}{Model} & CatAvg & SampAvg & Air.  & Bag   & Cap   & Car   & Chair & Ear.  & Guitar & Knife & Lamp  & Lap.  & Motor. & Mug   & Pistol & Rocket & Skate. & Table \\
    \midrule
    \multicolumn{2}{c}{\multirow{3}[2]{*}{\begin{sideways}\textbf{\footnotesize{Ful.Sup.}}\end{sideways}}} & PointNet\cite{Qi2017} & 80.4  & 83.7  & 83.4  & 78.7  & 82.5  & 74.9  & 89.6  & 73.0  & 91.5  & 85.9  & 80.8  & 95.3  & 65.2  & 93.0  & 81.2  & 57.9  & 72.8  & 80.6 \\
    \multicolumn{2}{c}{} & PointNet++\cite{qi2017pointnet++} & 81.9  & 85.1  & 82.4  & 79.0  & \textbf{87.7} & \textbf{77.3} & 90.8  & 71.8  & 91.0  & 85.9  & \textbf{83.7} & 95.3  & \textbf{71.6} & \textbf{94.1} & 81.3  & 58.7  & \textbf{76.4} & \textbf{82.6} \\
    \multicolumn{2}{c}{} & DGCNN\cite{wang2019dynamic} & \textbf{82.3} & \textbf{85.1} & \textbf{84.2} & \textbf{83.7} & 84.4  & 77.1  & \textbf{90.9} & \textbf{78.5} & \textbf{91.5} & \textbf{87.3} & 82.9  & \textbf{96.0} & 67.8  & 93.3  & \textbf{82.6} & \textbf{59.7} & 75.5  & 82.0 \\
    \midrule
    \multicolumn{2}{c}{\multirow{2}[1]{*}{\begin{sideways}\textbf{\footnotesize{Unsup.}}\end{sideways}}} & Kmeans & 39.4  & 39.6  & 36.3  & 34.0  & 49.7  & 18.0  & 48.0  & 37.5  & 47.3  & 75.6  & 42.0  & 69.7  & 16.6  & 30.3  & 43.3  & 33.1  & 17.4  & 31.7 \\
    \multicolumn{2}{c}{} & Ncut\cite{shi2000normalized}  & 43.5  & 43.2  & 41.0  & 38.0  & 53.4  & 20.0  & 52.1  & 41.1  & 52.1  & 83.5  & 46.1  & 77.5  & 18.0  & 33.5  & 48.0  & 36.5  & 19.6  & 35.0 \\
    \midrule
    \multirow{10}[3]{*}{\begin{sideways}\textbf{Weak Sup.}\end{sideways}} & \multirow{5}[1]{*}{\begin{sideways}1pt\end{sideways}} & $\Pi$ Model\cite{Laine_ICLR17} & 72.7	& 73.2	& 71.1	& \textbf{77.0}	& 76.1	& 59.7	& 85.3	& \textbf{68.0}	& 88.9	& 84.3	& 76.5	& \textbf{94.9}	& 44.6	& 88.7	& 74.2	& 45.1	& 67.4	& 60.9 \\
          &       & MT\cite{tarvainen2017mean}    & 68.6	& 72.2  & 71.6	& 60.0	& \textbf{79.3}	& 57.1	& 86.6	& 48.4	& 87.9	& 80.0	& 73.7	& 94.0	& 43.3	& 79.8	& 74.0	& 45.9	& 56.9	& 59.8 \\
          &       & Baseline & 72.2	& 72.6	& 74.3	& 75.9	& 79.0	& \textbf{64.2}	& 84.1	& 58.8	& 88.8	& 83.2	& 72.3	& 94.7	& 48.7	& 84.8	& 75.8	& \textbf{50.6}	& 60.3	& 59.5 \\
          &       & Ours  & \textbf{74.4}	& \textbf{75.5}	& \textbf{75.6}	& 74.4	& {79.2}	& 66.3	& \textbf{87.3}	& 63.3	& \textbf{89.4}	& \textbf{84.4}	& \textbf{78.7}	& 94.5	& \textbf{49.7}	& \textbf{90.3}	& \textbf{76.7}	& 47.8	& \textbf{71.0}	& \textbf{62.6} \\
\cmidrule{2-21}          & \multirow{5}[2]{*}{\begin{sideways}10\%\end{sideways}} & $\Pi$ Model\cite{Laine_ICLR17} & 79.2	& 83.8	& 80.0	& 82.3	& 78.7	& 74.9	& 89.8	& \textbf{76.8}	& 90.6	& 87.4	& \textbf{83.1}	& 95.8	& 50.7	& 87.8	& 77.9	& 55.2	& 74.3	& 82.7 \\
          &       & MT\cite{tarvainen2017mean}    & 76.8  & 81.7  & 78.0  & 76.3  & 78.1  & 64.4  & 87.6  & 67.2  & 88.7  & 85.5  & 79.0  & 94.3  & 63.3  & 90.8  & 78.2  & 50.7  & 67.5  & 78.5 \\
          &       & Baseline & 81.5	& 84.5	& 82.5	& 80.6	& \textbf{85.7}	& 76.4	& 90.0	& 76.6	& 89.7	& 87.1	& 82.6	& 95.6	& 63.3	& 93.6	& 79.7	& \textbf{63.2}	& 74.4	& 82.6 \\
          &       & Ours  & \textbf{81.7}	& \textbf{85.0}	& \textbf{83.1}	& \textbf{82.6}	& 80.8	& \textbf{77.7}	& \textbf{90.4}	& 77.3	& \textbf{90.9}	& \textbf{87.6}	& 82.9	& \textbf{95.8}	& \textbf{64.7}	& \textbf{93.9}	& \textbf{79.8}	& 61.9	& \textbf{74.9}	& \textbf{82.9} \\
    \bottomrule
    \end{tabular}%
    }
  \label{tab:ShapeNet}%
\end{table*}%

\vspace{-0.4cm}
\paragraph{S3DIS.} The results are presented in Tab.~\ref{tab:S3DIS_PartNet}. We make observations in a similar way to ShapeNet. First, the 1pt weak supervision provides strong results. The results of our proposed multi-task model is only $1\%$ lower than the fully supervised counterpart. Furthermore, the results 
of our method with only $10\%$ labelled points is even slightly superior than the fully supervision. Finally, the results of our method consistently outperform both unsupervised and alternative weakly supervised methods.

\vspace{-0.4cm}
\paragraph{PartNet.} For the PartNet dataset, we report the average mIoU in Tab.~\ref{tab:S3DIS_PartNet}. Details for each category is included in the supplementary. We also observe the same patterns from the results. The 1pt setting yields particularly strong results and our own variant outperforms all unsupervised and alternative weak supervision methods.

\begin{table*}[htbp]
        \begin{minipage}{0.79\textwidth}
  \centering
  \caption{\footnotesize{mIoU ($\%$) evaluations on S3DIS (Area 5) and PartNet datasets. We compared against fully supervised (Ful.Sup.), unsupervised (Unsup.) and alternative weakly supervised (Weak. Sup.) approaches.}}\vspace{-0.3cm}
  \setlength\tabcolsep{2pt} 
	  \resizebox{1\linewidth}{!}{
           \begin{tabular}{cclcccccccccccccc|cc}
    \toprule
          &       &       & \multicolumn{14}{c|}{S3DIS}                                                                                   & \multicolumn{2}{c}{PartNet} \\
\cmidrule(lr){4-17} \cmidrule(lr){18-19}    \multicolumn{2}{c}{Setting} & Model & \multicolumn{1}{l}{CatAvg} & \multicolumn{1}{l}{ceil.} & \multicolumn{1}{l}{floor} & \multicolumn{1}{l}{wall} & \multicolumn{1}{l}{beam} & \multicolumn{1}{l}{col.} & \multicolumn{1}{l}{win.} & \multicolumn{1}{l}{door} & \multicolumn{1}{l}{chair} & \multicolumn{1}{l}{table} & \multicolumn{1}{l}{book.} & \multicolumn{1}{l}{sofa} & \multicolumn{1}{l}{board} & \multicolumn{1}{l|}{clutter} & CatAvg & SampAvg \\
\midrule
    \multicolumn{2}{c}{\multirow{3}[1]{*}{\begin{sideways}\textbf{\footnotesize{Ful.Sup.}}\end{sideways}}} & PointNet & 41.1  & 88.8  & 97.3  & 69.8  & 0.1   & 3.9   & 46.3  & 10.8  & 52.6  & 58.9  & \textbf{40.3}  & 5.9   & 26.4  & 33.2  & 57.9  & 58.3 \\
    \multicolumn{2}{c}{} & PointNet++ & \textbf{47.8} & 90.3	& 95.6	& 69.3	& 0.1	& \textbf{13.8}	& 26.7	& \textbf{44.1}	& 64.3	& \textbf{70.0}	& 27.8	& \textbf{47.8}	& 30.8	& 38.1 & 65.5  & 67.1
 \\
    \multicolumn{2}{c}{} & DGCNN & 47.0  & \textbf{92.4}	& \textbf{97.6}	& \textbf{74.5}	& \textbf{0.5}	& 13.3	& \textbf{48.0}	& 23.7	& \textbf{65.4}	& 67.0	& 10.7	& 44.0	& \textbf{34.2}	& \textbf{40.0} & \textbf{65.6} & \textbf{67.2}
 \\
    \midrule
    \multicolumn{2}{c}{\multirow{2}[1]{*}{\begin{sideways}\textbf{\footnotesize{Unsup.}}\end{sideways}}} & Kmeans & 38.4  & 59.8  & 63.3  & 34.9  & 21.5  & \textbf{24.6} & 34.2  & 29.3  & 35.7  & 33.1  & 45.0  & 45.6  & 41.7  & 30.4  & 34.6  & 35.2 \\
    \multicolumn{2}{c}{} & Ncut  & \textbf{40.0} & \textbf{63.5} & \textbf{63.8} & \textbf{37.2} & \textbf{23.4} & \textbf{24.6} & \textbf{35.5} & \textbf{29.9} & \textbf{38.9} & \textbf{34.3} & \textbf{47.1} & \textbf{46.3} & \textbf{44.1} & \textbf{31.5} & \textbf{38.6} & \textbf{40.1} \\
    \midrule
    \multirow{10}[3]{*}{\begin{sideways}\textbf{Weak Sup.}\end{sideways}} & \multirow{5}[1]{*}{\begin{sideways}1pt\end{sideways}} & $\Pi$ Model & 44.3	& 89.1	& 97.0	& 71.5	& 0.0	& \textbf{3.6}	& 43.2	& 27.4	& 62.1	& 63.1	& 14.7	& \textbf{43.7}	& \textbf{24.0}	& 36.7 & 51.4 & 52.6 \\
          &       & MT    & 44.4	& 88.9	& 96.8	& 70.1	& \textbf{0.1}	& 3.0	& 44.3	& 28.8	& \textbf{63.6}	& 63.7	& 15.5	& \textbf{43.7}	& 23.0	& 35.8 & 52.9 & 53.6\\
          &       & Baseline & 44.0	& 89.8	& 96.7	& 71.5	& 0.0	& 3.0	& 43.2	& \textbf{32.8}	& 60.8	& 58.7	& 15.0	& 41.2	& 22.5	& 36.8 & 50.2 & 51.4  \\
          &       & Ours  & \textbf{44.5}	& \textbf{90.1}	& \textbf{97.1}	& \textbf{71.9}	& 0.0	& 1.9	& \textbf{47.2}	& 29.3	& 62.9	& \textbf{64.0}	& \textbf{15.9}	& 42.2	& 18.9	& \textbf{37.5} & \textbf{54.6} & \textbf{55.7}\\
\cmidrule{2-19}          & \multirow{5}[2]{*}{\begin{sideways}10\%\end{sideways}} & $\Pi$ Model & 46.3	& 91.8	& 97.1	& 73.8	& 0.0	& 5.1	& 42.0	& 19.6	& 66.7	& 67.2	& 19.1	& 47.9	& 30.6	& 41.3 & 64.1 & 64.7\\
          &       & MT    & 47.9	& 92.2	& 96.8	& 74.1	& 0.0	& 10.4	& 46.2	& 17.7	& 67.0	& 70.7	& \textbf{24.4}	& 50.2	& \textbf{30.7}	& 42.2 & 63.8 & 64.5\\
          &       & Baseline & 45.7	& 92.3	& \textbf{97.4}	& \textbf{75.4}	& 0.0	& \textbf{11.7}	& 47.2	& 22.9	& 65.3	& 66.7	& 11.7	& 43.6	& 17.8	& 41.5 & 63.1 & 63.9 \\
          &       & Ours  & \textbf{48.0}	& \textbf{90.9}	& 97.3	& 74.8	& 0.0	& 8.4	& \textbf{49.3}	& \textbf{27.3}	& \textbf{69.0}	& \textbf{71.7}	& 16.5	& \textbf{53.2}	& 23.3	& \textbf{42.8} & \textbf{64.5} & \textbf{64.9}\\
    \bottomrule
    \end{tabular}%
    }
  \label{tab:S3DIS_PartNet}%
          \end{minipage}
 \hfill
        \begin{minipage}{0.2\textwidth}
        \centering
            \caption{\footnotesize{Comparisons of different labelling strategies on ShapeNet segmentation. All numbers are in $\%$.}}
            \vspace{-0.1cm}
   \setlength\tabcolsep{2pt} 
	  \resizebox{1\textwidth}{!}{
    \begin{tabular}{lcc}
    \toprule
    Label Strat. & CatAvg & SampAvg \\
    \midrule
    Samp=10\% & \multirow{2}[2]{*}{70.37} & \multirow{2}[2]{*}{77.71} \\
    Pts=100\% &       &  \\
    \midrule
    Samp=20\% & \multirow{2}[2]{*}{72.19} & \multirow{2}[2]{*}{78.45} \\
    Pts=50\% &       &  \\
    \midrule
    Samp=50\% & \multirow{2}[2]{*}{74.29} & \multirow{2}[2]{*}{79.65} \\
    Pts=20\% &       &  \\
    \midrule
    Samp=80\% & \multirow{2}[2]{*}{76.15} & \multirow{2}[2]{*}{80.18} \\
    Pts=12.5\% &       &  \\
    \midrule
    Samp=100\% & \multirow{2}[2]{*}{77.71} & \multirow{2}[2]{*}{80.94} \\
    Pts=10\% &       &  \\
    \bottomrule
    \end{tabular}%
  \label{tab:PtsVsSamp}%
}
          \end{minipage}
            \vspace{-0.3cm}
\end{table*}%

\subsection{Qualitative Examples}

\begin{figure*}
\centering
\includegraphics[width=1.02\linewidth]{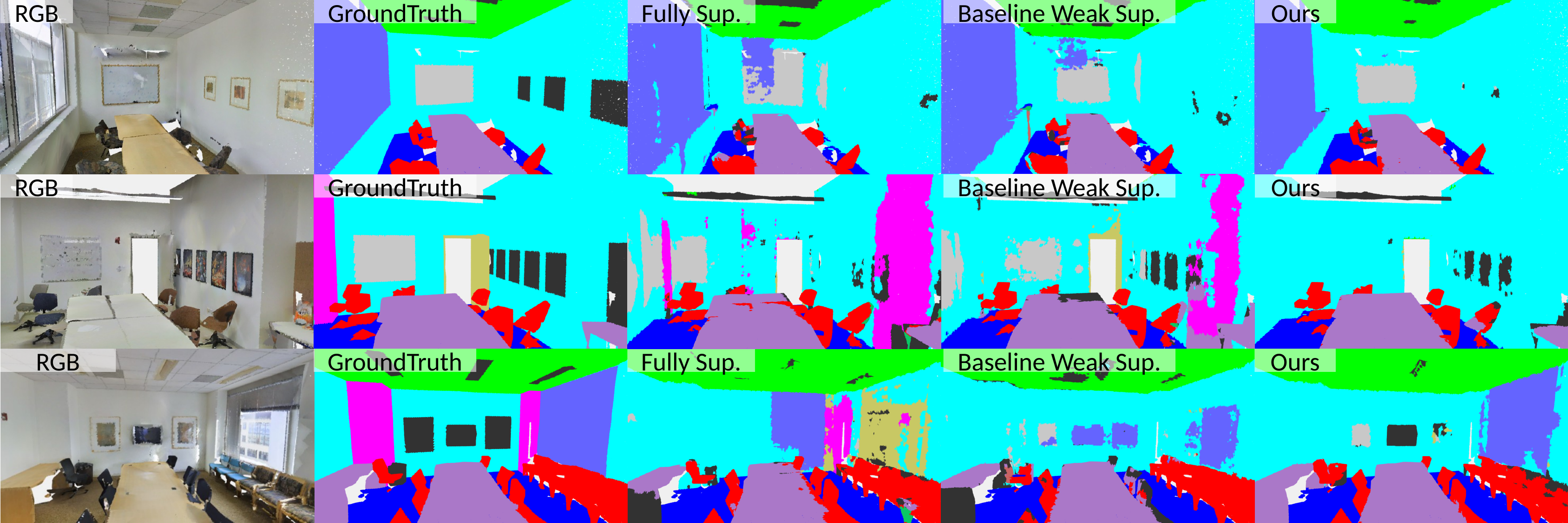}
\vspace{-0.3cm}
\caption{\footnotesize{Qualtitative examples for S3DIS dataset test area 5. $10\%$ labelled points are used to train the weak supervision models. }}\label{fig:S3DIS_Qualitative}
\vspace{-0.4cm}
\end{figure*}

\begin{figure}
\includegraphics[width=1\linewidth]{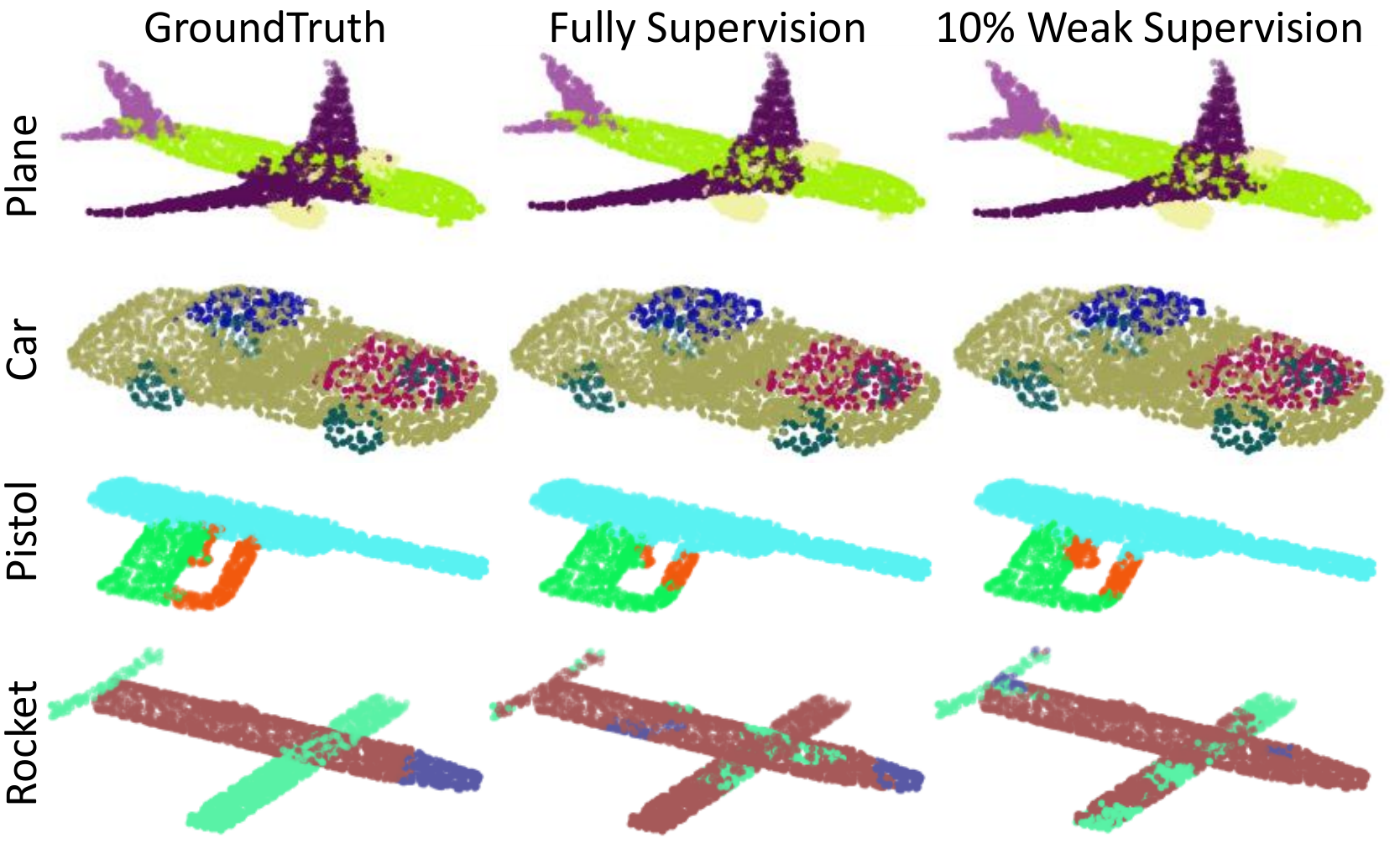}
\caption{\footnotesize{Qualitative examples for ShapeNet shape segmentation.}}\label{fig:ShapeNetExample}
\vspace{-0.4cm}
\end{figure}

We show qualitative examples of point cloud segmentation on all datasets and compare the segmentation quality. 
Firstly, we present the segmentation results on selected rooms from the S3DIS dataset in Fig.~\ref{fig:S3DIS_Qualitative}. From left to right we sequentially visualize the RGB view, ground-truth, fully supervised segmentation, weakly supervised baseline method and our final approach results. For weakly supervised methods, $10\%$ training points are assumed to be labelled. We observe accurate segmentation of majority and continuous objects, e.g. wall, floor, table, chair and window. In particular, our proposed method is able to improve the baseline results substantially by smoothing out the noisy areas. Nonetheless, we observe some mistakes of our method at the boundaries between different objects.  
The segmentation results on ShapeNet are shown in Fig.~\ref{fig:ShapeNetExample}. These examples again demonstrate the highly competitive performance by the weakly supervised approach. For both the plane and car categories, the results of the weak supervision are very close to the fully supervised ones.

\subsection{Label More Points Or More Samples}\label{sect:PtsVsSamp}
Given a fixed annotation budget, e.g. the total number of labelled points, there are different combinations of labelling strategies to balance the amount of labelled samples and the amount of labelled points within each sample. In this experiment, we control these two variables and validate on ShapeNet segmentation with the PointNet encoder for efficient evaluation. We first restrict the fixed budget to be $10\%$ of all training points. The labelling strategy is described by $x\%$ samples (Samp) each with $y\%$ labelled points (Pts) and $xy=1000$ to satisfy the restriction. We evaluate 5 combinations and present the results in Tab.~\ref{tab:PtsVsSamp}. The consistent improvement of mIoU with $x\%$ from $10\%$ to $100\%$ suggests that, given fixed total annotation budget, it is better to extensively label more samples each with fewer labelled points than intensively labelling a fraction of the dataset.

\section{Ablation Study}
\paragraph{Importance of Individual Components.}
We analyze the importance of the proposed additional losses and inference label propagation. Different combinations of the losses are evaluated on all datasets with the 1pt annotation scheme. The results are presented in Tab.~\ref{tab:AblLoss}. We observe that the Siamese self-supervision introduces the most advantage for S3DIS. This is because S3DIS is a real dataset, where the orientations and layouts of objects are diverse, and the augmentation and consistency constraints increase the robustness of model. 
In contrast, the pose of test shapes are always fixed for the other two datasets, and thus they benefit less from Siamese augmentation. We also compare against the use of only data augmentation (last row), and the results suggest it is better to have the consistency constraints on unlabelled points. 
The results are also further improved with the multi-instance loss for inexact branch. Finally, the smooth constraint at both training (Smo.) and inference (TeLP) stages consistently bring additional advantage to the whole architecture.

\vspace{-0.4cm}
\paragraph{Compatibility with Encoder Network.}\label{sect:EncoderNet}
We further examine the compatibility of the proposed losses with different encoder networks. In particular, we investigate the performance with PointNet and DGCNN as the encoder network. The results are shown in Tab.~\ref{tab:AblLoss} and it is clear that both networks exhibit same patterns.

\begin{table}[htbp]
  \centering
 \caption{\small{Ablation study on the impact of individual losses and inference label propagation and the compatibility with alternative encoder networks.}}\vspace{-0.2cm}
   \setlength\tabcolsep{2pt} 
  \resizebox{0.9\linewidth}{!}{
      \begin{tabular}{cccc|ccc|ccc}
    \toprule
    \multicolumn{4}{c|}{Components} & \multicolumn{3}{c|}{PointNet} & \multicolumn{3}{c}{DGCNN} \\
    \midrule
    MIL   & Siam. & Smo.  & TeLP  & ShapeNet & PartNet & S3DIS & ShapeNet & PartNet & S3DIS \\    
    \midrule
          &       &       &       & 65.2  & 49.7  & 36.8  & 72.2  & 50.2  & 44.0 \\
          & \checkmark     &       &       & 66.0  & 50.3  & 41.9  & 73.1  & 51.5  & 44.3 \\
    \checkmark     & \checkmark     &       &       & 69.0  & 52.1  & 42.2  & 73.4  & 52.9  & 44.4 \\
    \checkmark     & \checkmark     & \checkmark     &       & 69.6  & 52.5  & 43.0  & 73.8  & 53.6  & 44.2 \\
    \checkmark     & \checkmark     & \checkmark     & \checkmark     & \textbf{70.2} & \textbf{52.8} & \textbf{43.1} & \textbf{74.4} & \textbf{54.6} & \textbf{44.5} \\
    \midrule
    \multicolumn{4}{c|}{Data Augmentation} & 65.3  & 49.9  & 38.9  & 73.0  & 52.7  & 43.2 \\
    \bottomrule
    \end{tabular}%
    }
  \label{tab:AblLoss}%
  \vspace{-0.3cm}
\end{table}%

\begin{figure}
\centering
\subfloat[ShapeNet]{\includegraphics[width=0.34\linewidth]{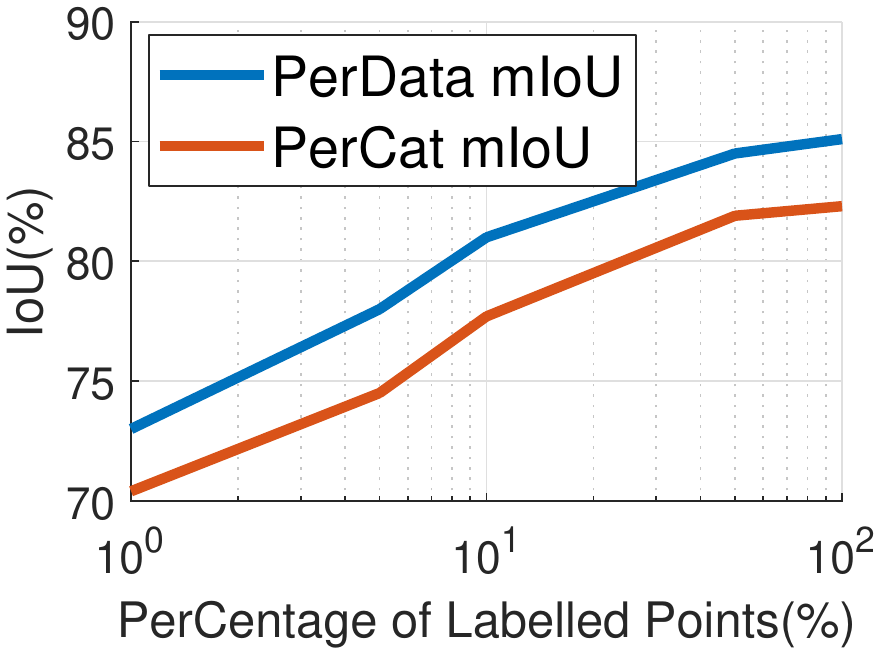}}
\subfloat[PartNet]{\includegraphics[width=0.34\linewidth]{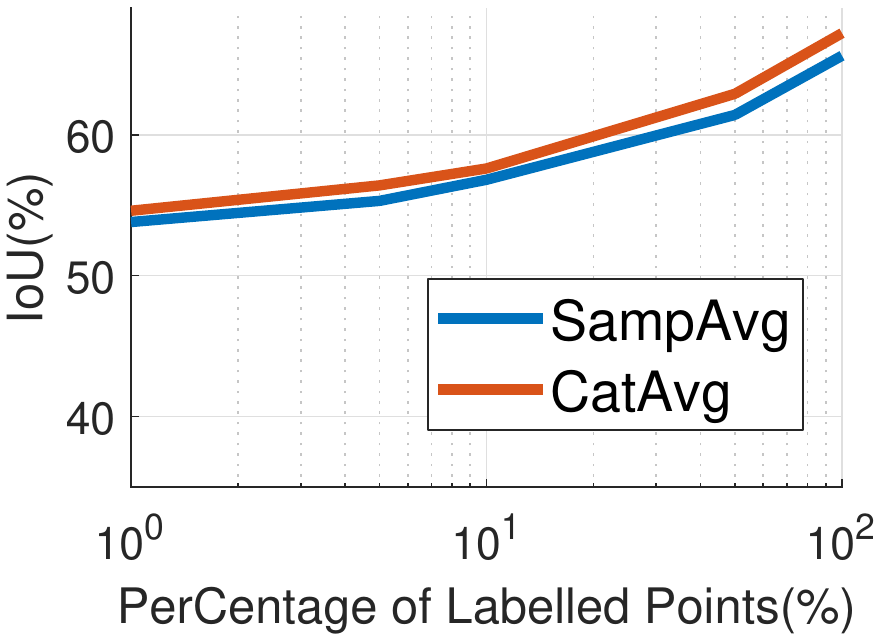}}
\subfloat[S3DIS]{\includegraphics[width=0.34\linewidth]{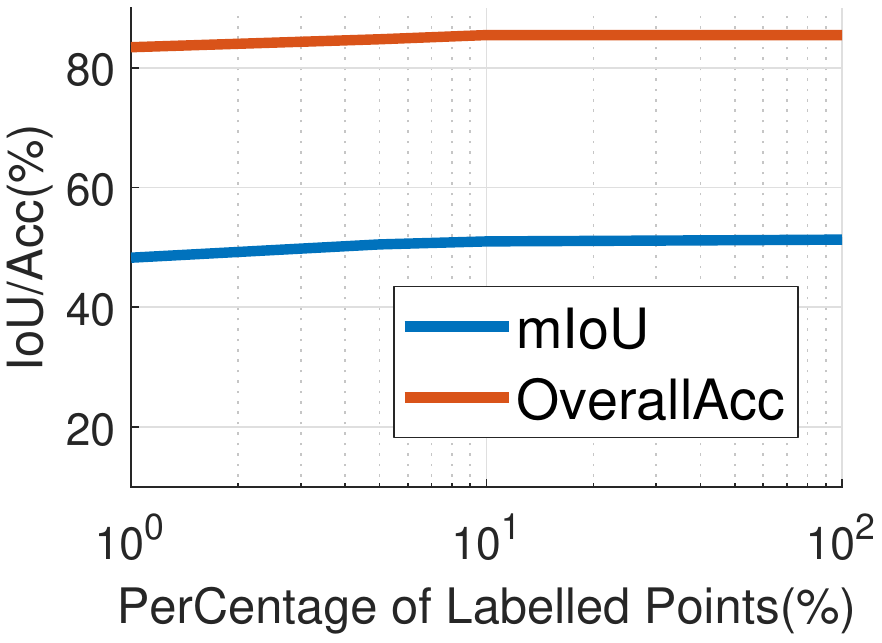}}
\vspace{-0.3cm}
\caption{\small{The impact of amount of labelled points for all three datasets.}}\label{fig:AmtLabelData}
\vspace{-0.3cm}
\end{figure}

\vspace{-0.4cm}
\paragraph{Amount of Labelled Data.}
As suggested by previous study, the amount of labelled data has an significant impact on the point cloud segmentation performance. In this section, we investigate this relation by varying the amount of labelled points. In particular, we control the percentage of labelled points to be from $1\%$ to $100\%$ (full supervision) with the baseline weak supervision method. The results are presented in Fig.~\ref{fig:AmtLabelData}. We observe that the performance on all datasets approaches the full supervision after $10\%$ labelled points.

\begin{figure}
\includegraphics[width=1.05\linewidth]{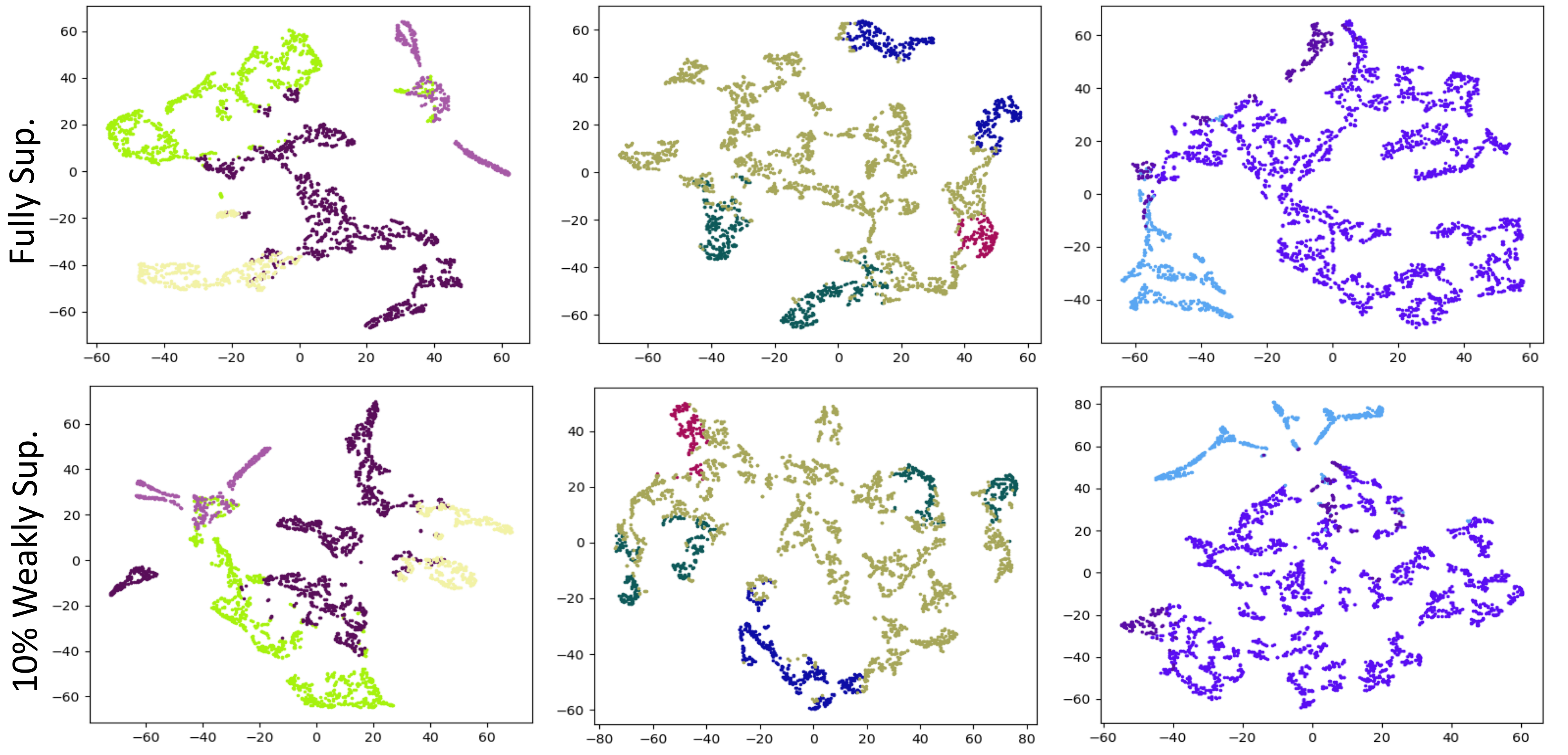}
\vspace{-0.5cm}
\caption{\small{T-SNE visualization of point embeddings in 2D space.}}\label{fig:PtEmbedTSNE}
\vspace{-0.5cm}
\end{figure}

\vspace{-0.4cm}
\paragraph{Point Feature Embedding.}
We visualize the point cloud feature embedding to further understand why weak supervision leads to competitive performance. We first project the feature before the last layer into 2D space via T-SNE \cite{maaten2008visualizing} for both full supervision and $10\%$ weak supervision. The projected point embeddings are visualized in Fig.~\ref{fig:PtEmbedTSNE}. We observe similar feature embedding patterns. This again demonstrates a few labelled points can yield very competitive performance.

\section{Conclusion}
  \vspace{-0.2cm}
In this paper, we made a discovery that only a few labelled points is needed for existing point cloud encoder networks to produce very competitive performance
for the point cloud segmentation task. We provide analysis from a statistical point of view and gave insights into the annotation strategy under fixed labelling budget. Furthermore, we proposed three additional training losses, i.e. inexact supervision, Siamese self-supervision and spatial and color smoothness to further regularize the model. Experiments are carried out on three public datasets to validate the efficacy of our proposed methods. In particular, the results are comparable with full supervision with 10 $\times$ fewer labelled points. 

\vspace{-0.4cm}
\paragraph{Acknowledgement.} This work was partially supported by the Singapore MOE Tier 1 grant R-252-000-A65-114.

{\small
\bibliographystyle{ieee_fullname}
\bibliography{references}
}

\appendix

\noindent\textbf{\large
{Appendix}}

\section{Compatibility with Alternative Encoder Networks}
We further evaluate an additional the state-of-the-art encoder network with the proposed weakly supervised strategy. Specifically, PointNet++ is evaluated on the ShapeNet dataset. The fully supervised setting (FullSup), 1 point per category labelling (1pt WeakSup) and 10$\%$ labelling (10$\%$ WeakSup) with our final  model are compared. The results in Tab.~\ref{tab:Encoder} clearly demonstrate that with very few annotations, shape segmentation is still very robust with different encoder networks.

\section{Additional Details on Datasets}
We present more details on the weakly supervised segmentation experiments on PartNet in Tab.~\ref{tab:PartNet}.


\section{Additional Qualitative Examples}
More qualitative examples on S3DIS and ShapeNet are presented here. We show the following for S3DIS in Fig. 1:  RGB
view, ground-truth segmentation (GT View), fully supervised
segmentation (FullSup. Seg.), baseline weakly supervised
method with $10\%$ labelled points ($10\%$ Baseline
WeakSup. Seg.), our final multi-task weakly supervised
method with $10\%$ points labelled ($10\%$ OurWeakSup. Seg.),
and our final multi-task weakly supervised method with 1
labelled point per category (1pt Our WeakSup. Seg.). Fig.1 shows 9 selected rooms in Area 5 of the S3DIS dataset. In these results, we observe consistent improvement of our method over baseline method. Moreover, the $10\%$ weak supervision results are even comparable to the fully supervised one and the 1pt weak supervision results is also surprisingly good. We further visualize additional segmentation results on the ShapeNet dataset with both 1pt and 10$\%$ weak supervision. The gap between weak supervision and full supervision is even smaller on the shape segmentation task.


\begin{table*}[!ht]
  \centering
  \caption{Evaluation of alternative encoder network on ShapeNet dataset.}
  \setlength\tabcolsep{2pt} 
  \resizebox{0.99\linewidth}{!}{
    \begin{tabular}{clrrrrrrrrrrrrrrrrrr}
    \toprule
    \multirow{2}[4]{*}{Encoder Net} & \multicolumn{1}{c}{\multirow{2}[4]{*}{Setting}} & \multicolumn{18}{c}{ShapeNet} \\
\cmidrule{3-20}          &       & \multicolumn{1}{c}{CatAvg} & \multicolumn{1}{c}{SampAvg} & \multicolumn{1}{c}{Air.} & \multicolumn{1}{c}{Bag} & \multicolumn{1}{c}{Cap} & \multicolumn{1}{c}{Car} & \multicolumn{1}{c}{Chair} & \multicolumn{1}{c}{Ear.} & \multicolumn{1}{c}{Guitar} & \multicolumn{1}{c}{Knife} & \multicolumn{1}{c}{Lamp} & \multicolumn{1}{c}{Lap.} & \multicolumn{1}{c}{Motor.} & \multicolumn{1}{c}{Mug} & \multicolumn{1}{c}{Pistol} & \multicolumn{1}{c}{Rocket} & \multicolumn{1}{c}{Skate.} & \multicolumn{1}{c}{Table} \\
    \midrule
    \multirow{3}[2]{*}{PointNet++} & FullSup & \multicolumn{1}{c}{81.87} & \multicolumn{1}{c}{84.89} & \multicolumn{1}{c}{82.74} & \multicolumn{1}{c}{81.19} & \multicolumn{1}{c}{87.84} & \multicolumn{1}{c}{78.11} & \multicolumn{1}{c}{90.71} & \multicolumn{1}{c}{73.40} & \multicolumn{1}{c}{90.93} & \multicolumn{1}{c}{86.04} & \multicolumn{1}{c}{83.36} & \multicolumn{1}{c}{95.07} & \multicolumn{1}{c}{72.38} & \multicolumn{1}{c}{94.96} & \multicolumn{1}{c}{80.03} & \multicolumn{1}{c}{55.13} & \multicolumn{1}{c}{76.05} & \multicolumn{1}{c}{81.98} \\
          & 1pt WeakSup & \multicolumn{1}{c}{80.82} & \multicolumn{1}{c}{84.19} & \multicolumn{1}{c}{80.86} & \multicolumn{1}{c}{76.90} & \multicolumn{1}{c}{86.94} & \multicolumn{1}{c}{75.57} & \multicolumn{1}{c}{90.35} & \multicolumn{1}{c}{74.00} & \multicolumn{1}{c}{90.34} & \multicolumn{1}{c}{86.05} & \multicolumn{1}{c}{83.66} & \multicolumn{1}{c}{95.12} & \multicolumn{1}{c}{66.97} & \multicolumn{1}{c}{93.22} & \multicolumn{1}{c}{79.20} & \multicolumn{1}{c}{57.93} & \multicolumn{1}{c}{74.30} & \multicolumn{1}{c}{81.68} \\
          & 10\% WeakSup & \multicolumn{1}{c}{81.27} & \multicolumn{1}{c}{84.70} & \multicolumn{1}{c}{82.36} & \multicolumn{1}{c}{76.55} & \multicolumn{1}{c}{86.82} & \multicolumn{1}{c}{77.48} & \multicolumn{1}{c}{90.52} & \multicolumn{1}{c}{73.01} & \multicolumn{1}{c}{91.16} & \multicolumn{1}{c}{85.35} & \multicolumn{1}{c}{83.07} & \multicolumn{1}{c}{95.34} & \multicolumn{1}{c}{69.97} & \multicolumn{1}{c}{94.88} & \multicolumn{1}{c}{80.04} & \multicolumn{1}{c}{57.03} & \multicolumn{1}{c}{74.59} & \multicolumn{1}{c}{82.14} \\
    \bottomrule
    \end{tabular}%
    }
  \label{tab:Encoder}%
\end{table*}%

\begin{table*}[htbp]
  \centering
  \caption{Detailed results on PartNet dataset.}
  \setlength\tabcolsep{2pt} 
  \resizebox{0.99\linewidth}{!}{
    \begin{tabular}{cclccccccccccccccccccccccccc}
    \toprule
    \multicolumn{2}{c}{Setting} & \multicolumn{1}{c}{Model} & CatAvg & Bag   & Bed   & Bott. & Bowl  & Chair & Clock & Dish. & Disp. & Door  & Ear.  & Fauc. & Hat   & Key   & Knife & Lamp  & Lap.  & Micro. & Mug   & Frid. & Scis. & Stora. & Table & Trash. & Vase \\
    \midrule
    \multicolumn{2}{c}{\multirow{3}[2]{*}{\begin{sideways}\textbf{Ful.Sup.}\end{sideways}}} & PointNet & 57.9  & 42.5  & 32.0  & 33.8  & 58.0  & 64.6  & 33.2  & 76.0  & 86.8  & 64.4  & 53.2  & 58.6  & 55.9  & 65.6  & 62.2  & 29.7  & 96.5  & 49.4  & 80.0  & 49.6  & 86.4  & 51.9  & 50.5  & 55.2  & 54.7 \\
    \multicolumn{2}{c}{} & PointNet++ & 65.5  & 59.7  & 51.8  & \textbf{53.2} & \textbf{67.3} & 68.0  & \textbf{48.0} & \textbf{80.6} & 89.7  & 59.3  & \textbf{68.5} & \textbf{64.7} & 62.4  & 62.2  & 64.9  & \textbf{39.0} & \textbf{96.6} & 55.7  & \textbf{83.9} & 51.8  & 87.4  & 58.0  & \textbf{69.5} & 64.3  & 64.4 \\
    \multicolumn{2}{c}{} & DGCNN & \textbf{65.6} & \textbf{53.3} & \textbf{58.6} & 48.9  & 66.9  & \textbf{69.1} & 35.8  & 75.2  & \textbf{91.2} & \textbf{68.5} & 59.3  & 62.6  & \textbf{63.7} & \textbf{69.5} & \textbf{71.8} & 38.5  & 95.7  & \textbf{57.6} & 83.3  & \textbf{53.7} & \textbf{89.7} & \textbf{62.6} & 65.3  & \textbf{67.8} & \textbf{66.8} \\
    \midrule
    \multirow{4}[4]{*}{\begin{sideways}\textbf{Weak Sup.}\end{sideways}} & \multirow{2}[2]{*}{\begin{sideways}1pt\end{sideways}} & Baseline & 50.2 & 24.4 & 30.1 & 20.5 & 38.0 & 65.9 & 35.3 & 64.9 & \textbf{84.3} & \textbf{52.6} & 36.7 & 47.1 & 47.9 & 52.2 & 55.2 & 34.1 & 92.4 & 49.3 & 59.5 & 49.6 & 80.1 & 44.6 & 49.8 & 40.4 & 49.5 \\
          &       & Ours & \textbf{54.6} & \textbf{28.4} & \textbf{30.8} & \textbf{26.0} & \textbf{54.3} & \textbf{66.4} & \textbf{37.7} & \textbf{66.3} & 81.0 & 51.7 & \textbf{44.4} & \textbf{51.2} & \textbf{55.2} & \textbf{56.2} & \textbf{63.1} & \textbf{37.6} & \textbf{93.5} & \textbf{49.7} & \textbf{73.5} & \textbf{50.6} & \textbf{83.6} & \textbf{46.8} & \textbf{61.1} & \textbf{44.1} & \textbf{56.8} \\
\cmidrule{2-28}          & \multirow{2}[2]{*}{\begin{sideways}10\%\end{sideways}} & Baseline & 63.2 & 54.4 & 56.8 & 44.1 & \textbf{57.6} & 67.2 & 41.3 & \textbf{70.0} & \textbf{91.3} & \textbf{61.8} & \textbf{65.8} & 57.2 & \textbf{64.2} & 64.2 & 66.7 & \textbf{37.9} & 94.9 & 49.1 & 80.2 & \textbf{49.6} & 84.1 & 59.3 & 69.7 & 66.7 & 63.0 \\
          &       & Ours & \textbf{64.5} & \textbf{47.3} & \textbf{55.5} & \textbf{64.7} & 56.2 & \textbf{69.1} & \textbf{44.3} & 68.3 & 91.1 & 61.3 & 62.8 & \textbf{65.2} & 63.0 & \textbf{64.6} & \textbf{67.9} & 37.8 & \textbf{95.5} & \textbf{50.1} & \textbf{82.7} & \textbf{49.6} & \textbf{85.8} & \textbf{59.5} & \textbf{71.2} & \textbf{67.7} & \textbf{65.9} \\
    \bottomrule
    \end{tabular}%
    }
  \label{tab:PartNet}%
\end{table*}%

\begin{figure*}[!htb]

\caption{Additional examples comparing full supervision and weak supervision for S3DIS semantic segmentation.}\label{fig:S3DIS}


\subfloat[Area5\_conferenceRoom\_2]{\includegraphics[width=1\linewidth]{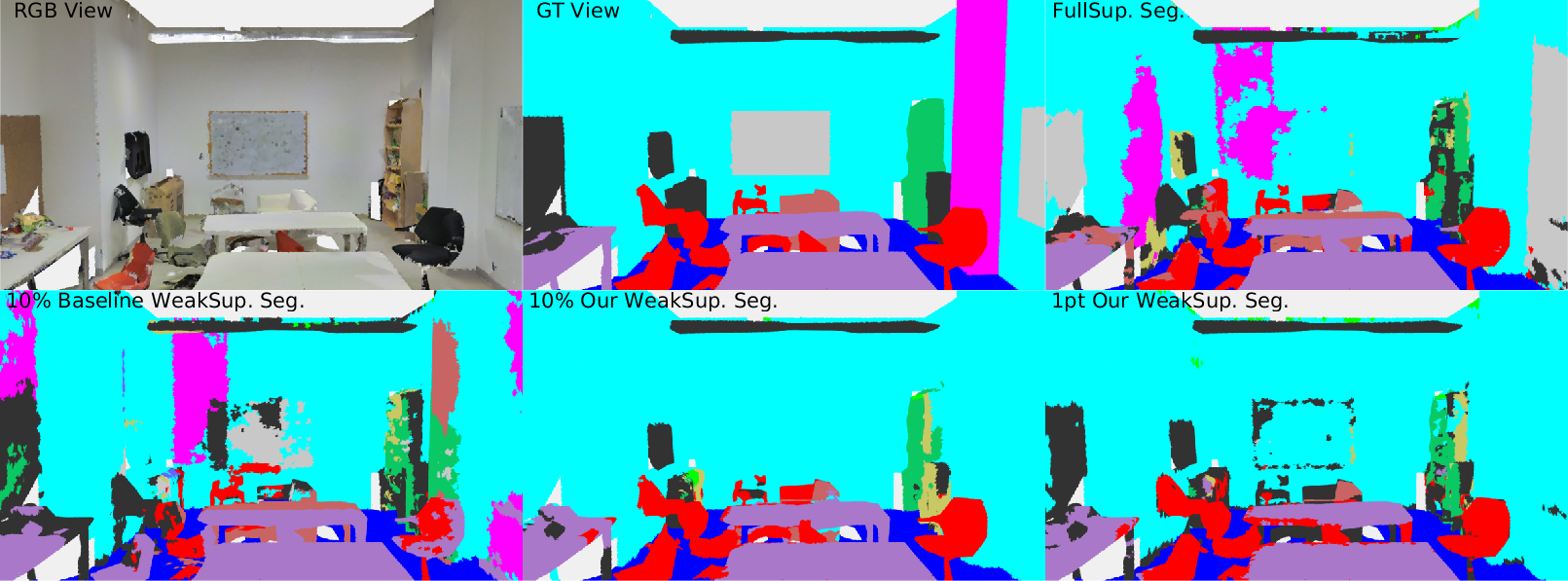}}

\subfloat[Area5\_conferenceRoom\_3]{\includegraphics[width=1\linewidth]{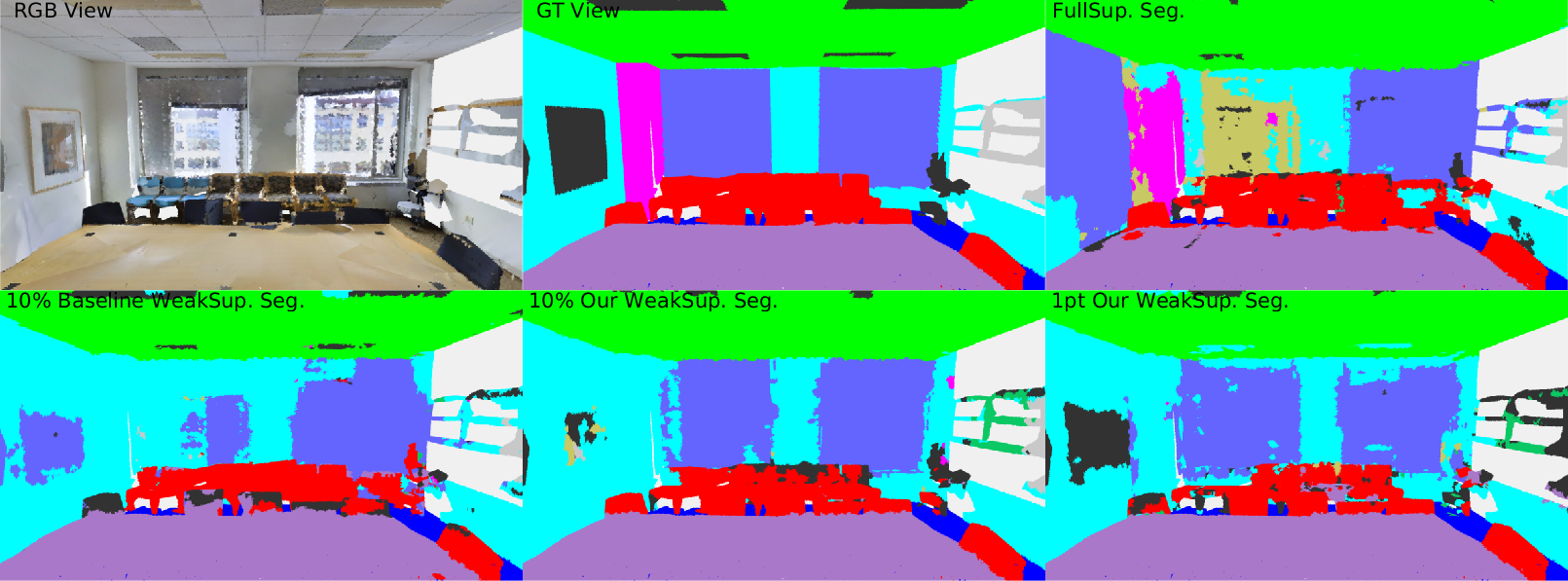}}

\end{figure*}
\begin{figure*}\ContinuedFloat

\subfloat[Area5\_hallway\_1]{\includegraphics[width=1\linewidth]{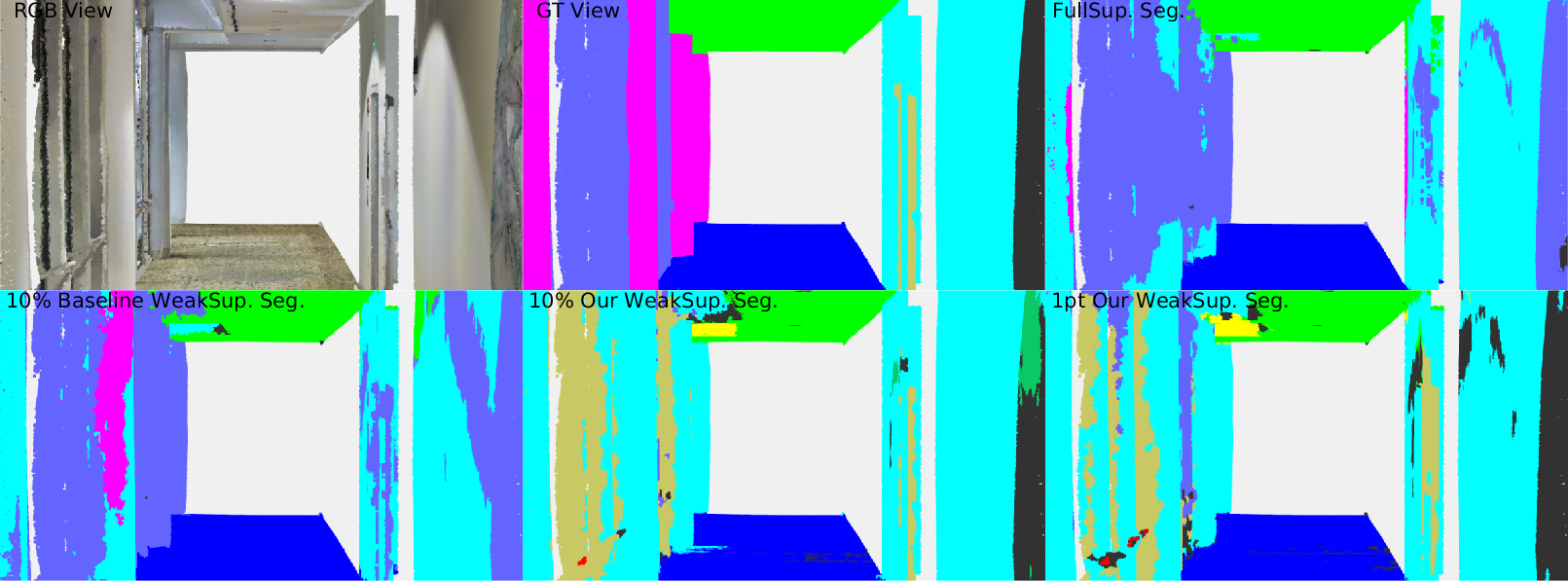}}

\subfloat[Area5\_lobby\_1]{\includegraphics[width=1\linewidth]{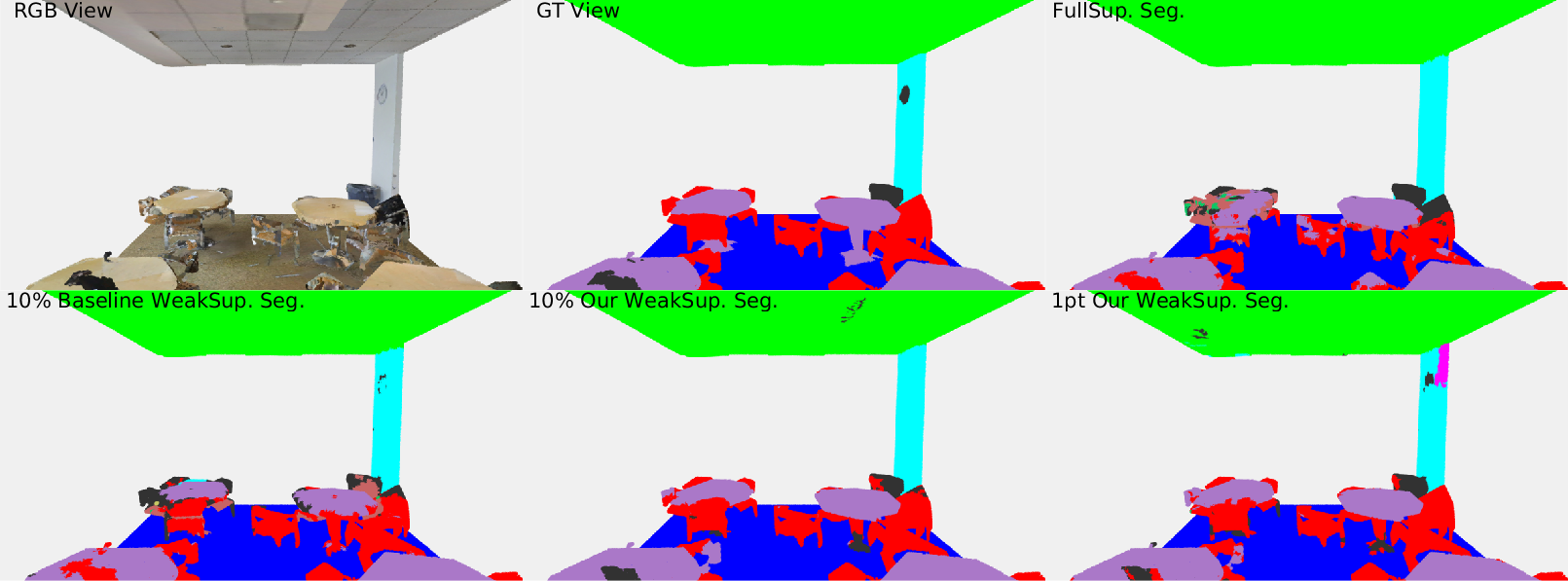}}

\subfloat[Area5\_office\_1]{\includegraphics[width=1\linewidth]{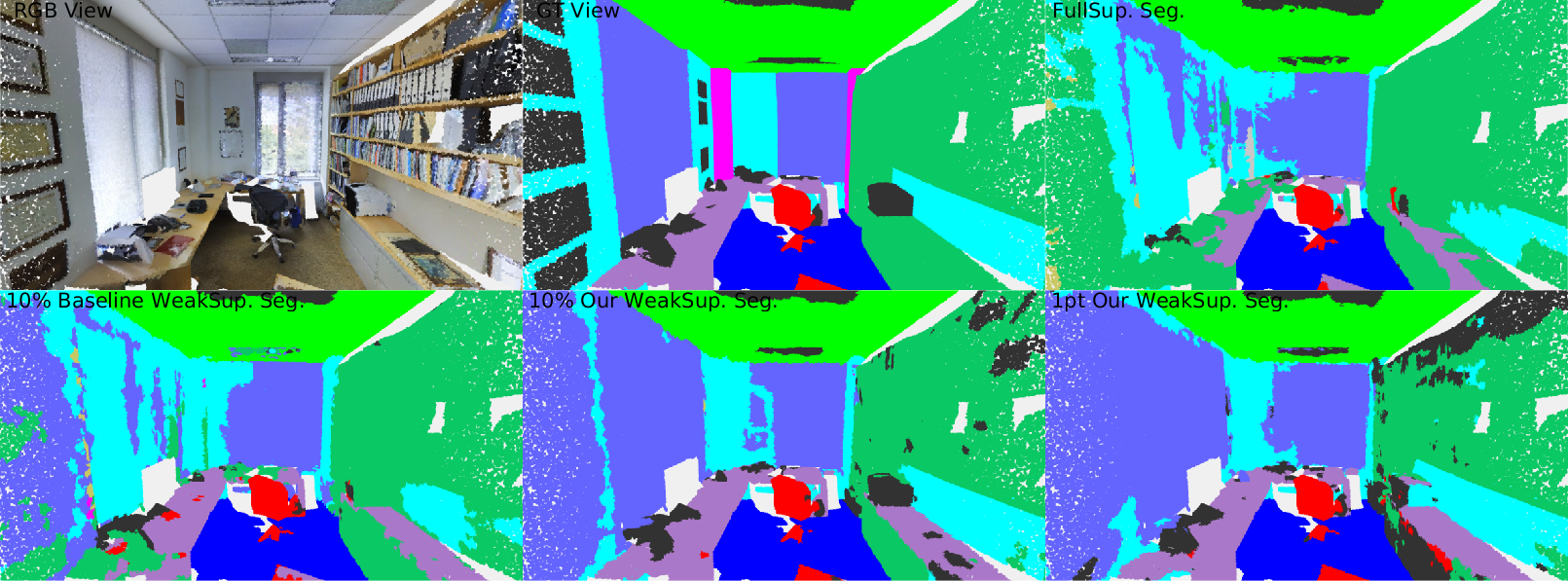}}

\end{figure*}
\begin{figure*}\ContinuedFloat

\subfloat[Area5\_office\_10]{\includegraphics[width=1\linewidth]{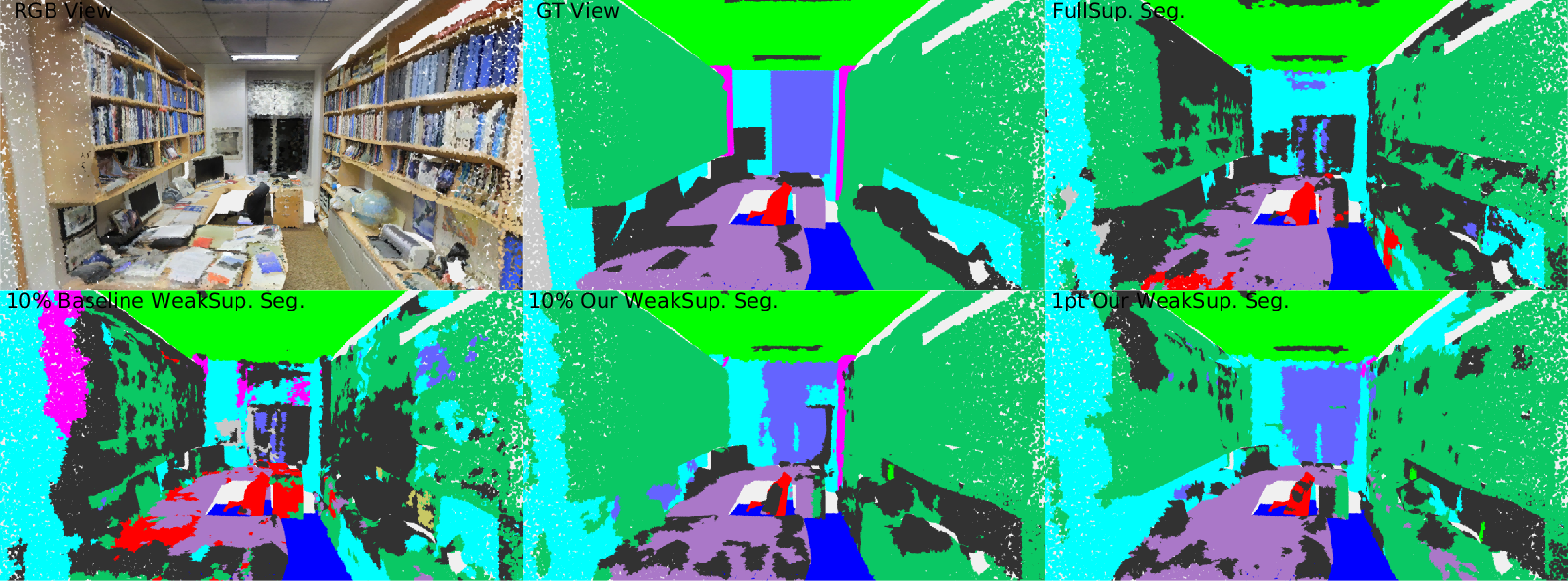}}

\subfloat[Area5\_pantry\_1]{\includegraphics[width=1\linewidth]{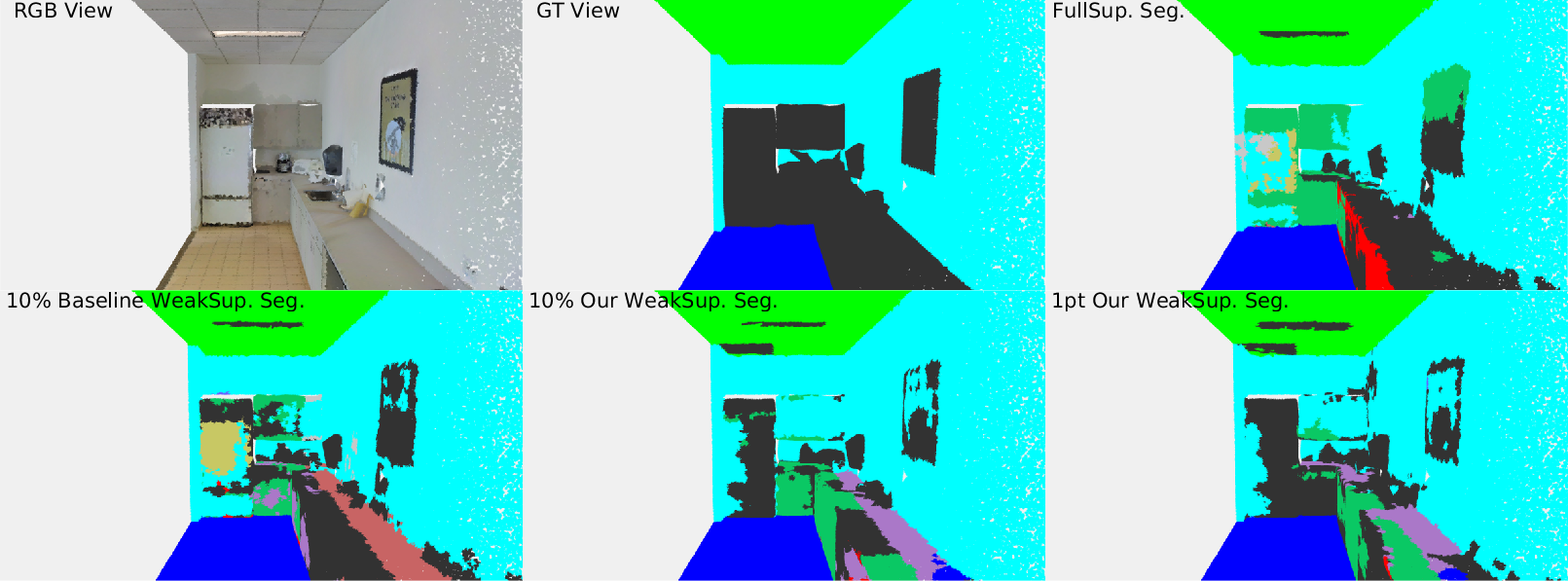}}

\subfloat[Area5\_storage\_2]{\includegraphics[width=1\linewidth]{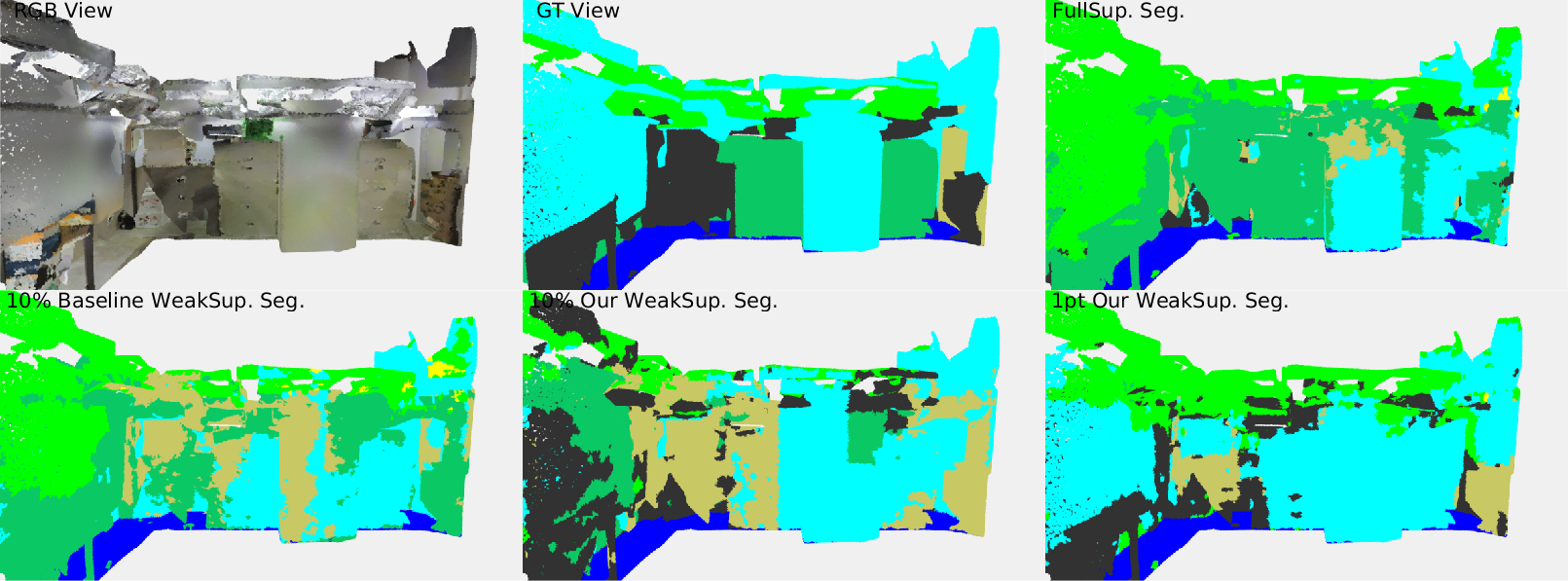}}

\end{figure*}

\begin{figure*}
\caption{Additional examples comparing full supervision and weak supervision for ShapeNet part segmentation.}
\subfloat{\includegraphics[width=1\linewidth]{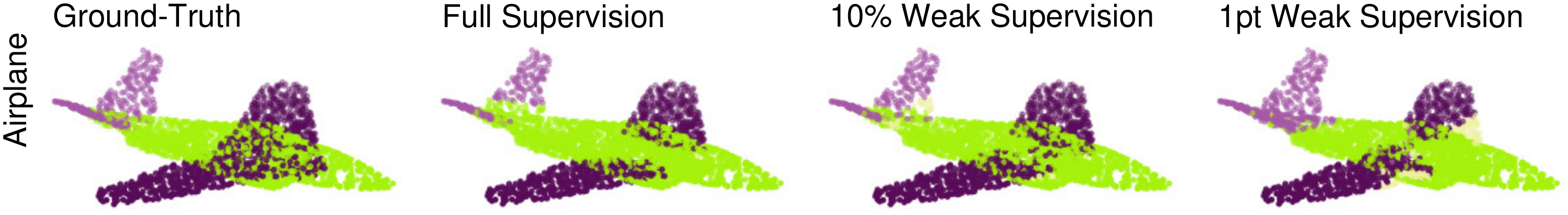}}\\
\subfloat{\includegraphics[width=1\linewidth]{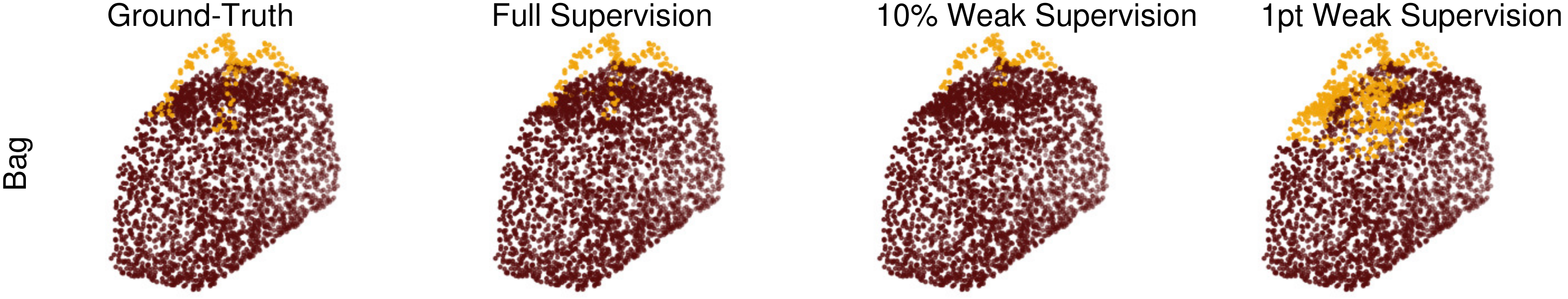}}\\
\subfloat{\includegraphics[width=1\linewidth]{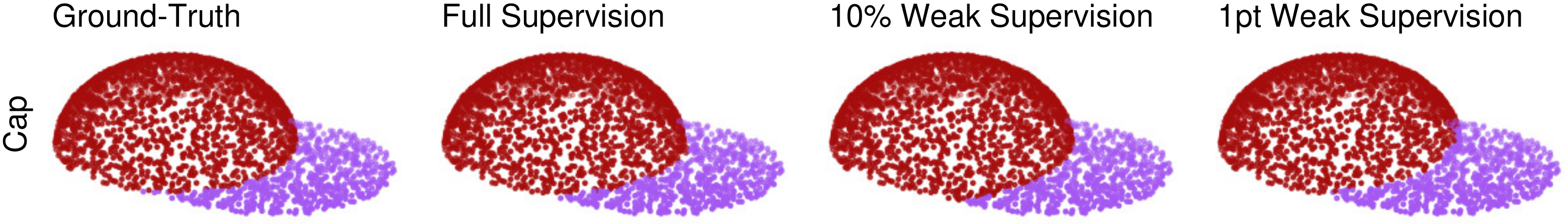}}\\
\subfloat{\includegraphics[width=1\linewidth]{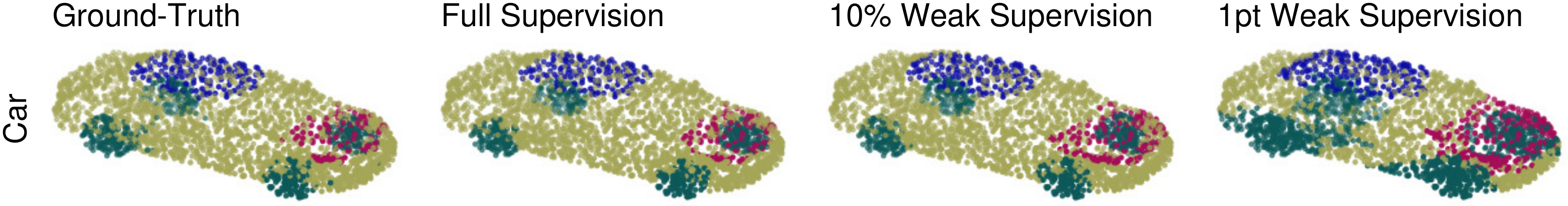}}\\
\subfloat{\includegraphics[width=1\linewidth]{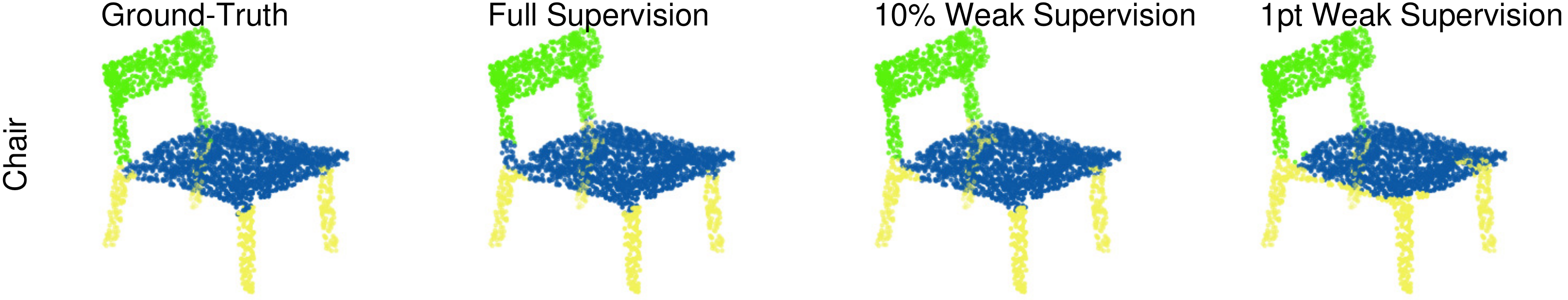}}\\
\subfloat{\includegraphics[width=1\linewidth]{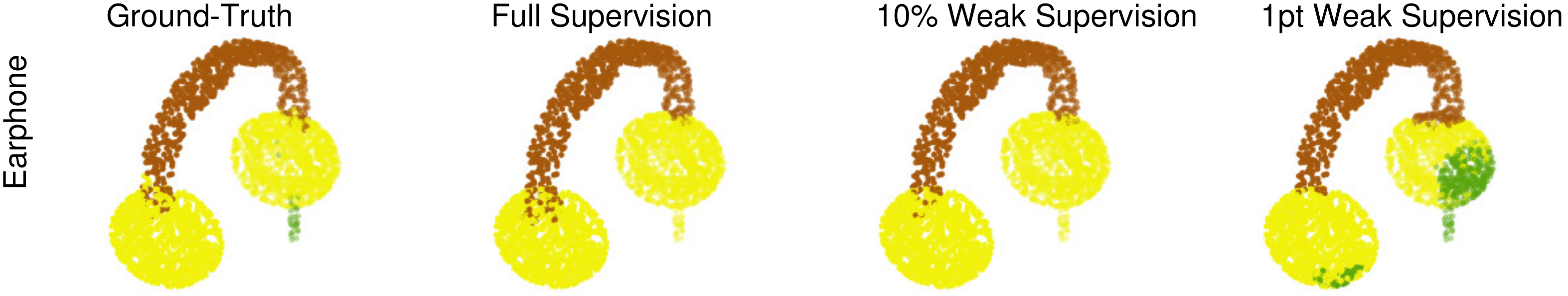}}\\
\subfloat{\includegraphics[width=1\linewidth]{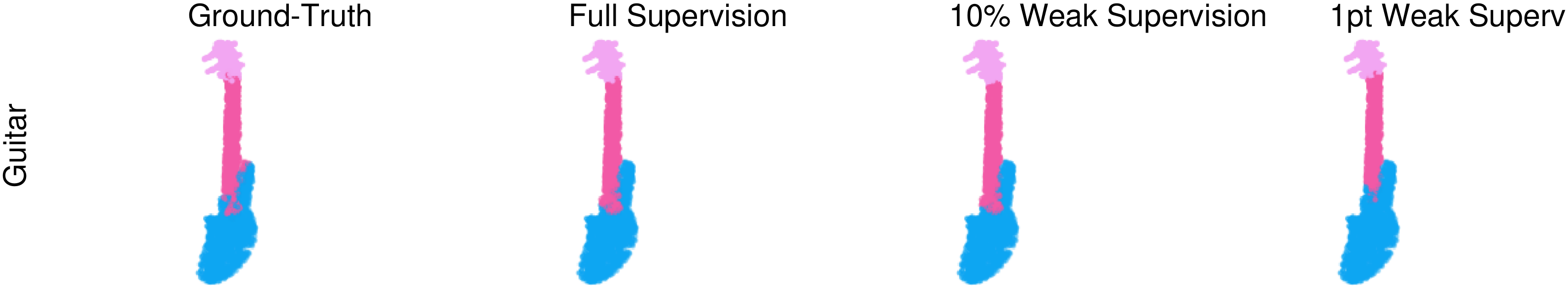}}\\
\end{figure*}

\begin{figure*}\ContinuedFloat
\subfloat{\includegraphics[width=1\linewidth]{./Figure/ShapeNet_Qualitative/Supp/Shapes/instance-1253_cat-Guitar.pdf}}\\
\subfloat{\includegraphics[width=1\linewidth]{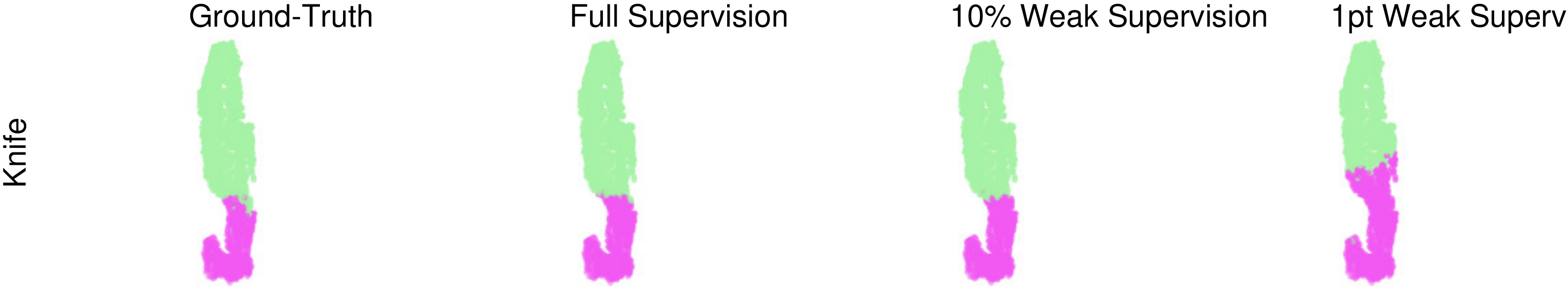}}\\
\subfloat{\includegraphics[width=1\linewidth]{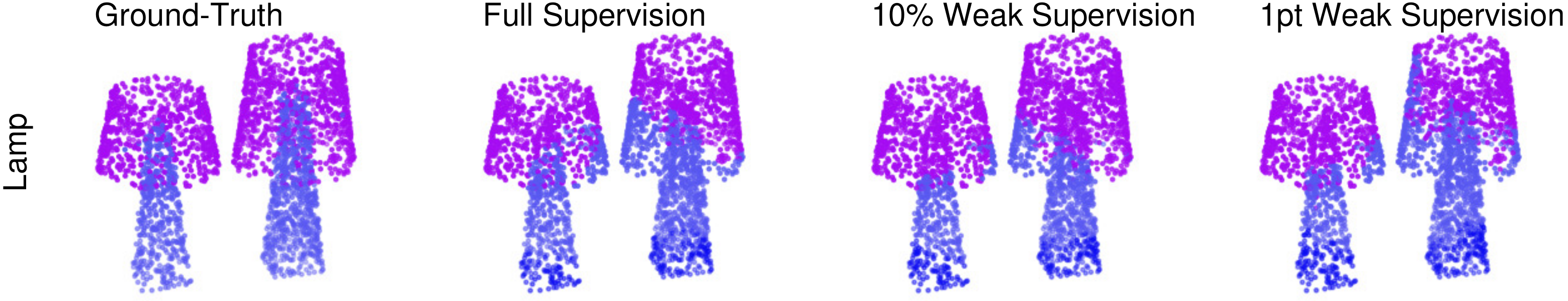}}\\
\subfloat{\includegraphics[width=1\linewidth]{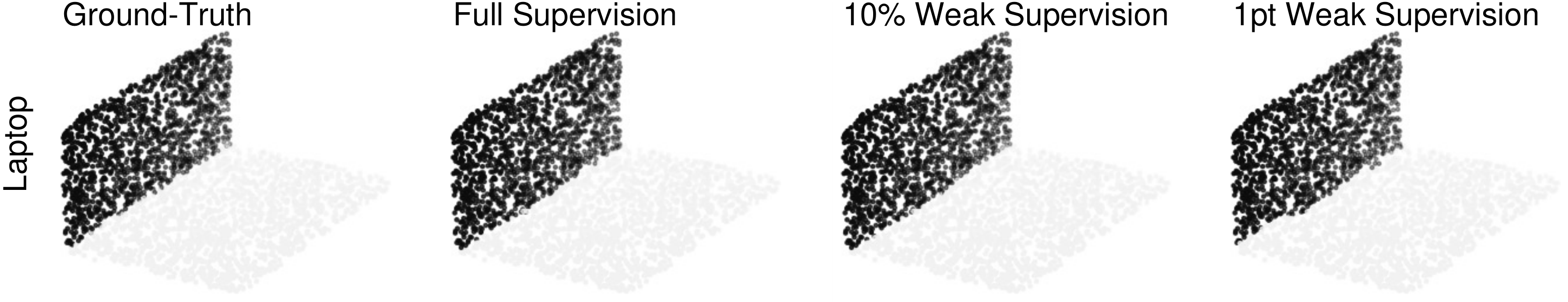}}\\
\subfloat{\includegraphics[width=1\linewidth]{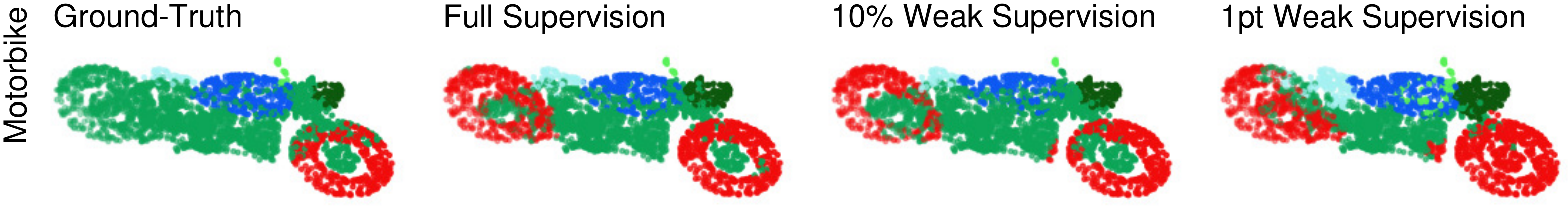}}\\
\subfloat{\includegraphics[width=1\linewidth]{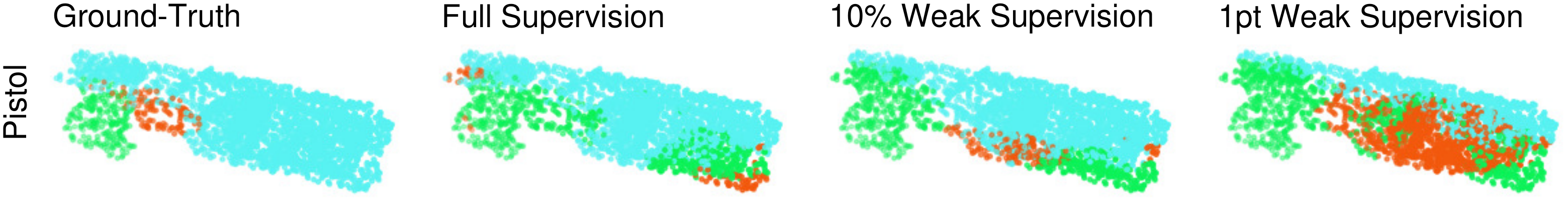}}\\
\end{figure*}


\end{document}